\documentclass[lettersize,journal]{IEEEtran}
\usepackage{amsmath,amsfonts}
\usepackage{algorithmic}
\usepackage{algorithm}
\usepackage{array}
\usepackage[caption=false,font=normalsize,labelfont=sf,textfont=sf]{subfig}
\usepackage{textcomp}
\usepackage{stfloats}
\usepackage{url}
\usepackage[inkscapelatex=false]{svg}
\usepackage{verbatim}
\usepackage{graphicx}
\usepackage{cite}
\usepackage{xcolor}
\usepackage[table]{xcolor}
\usepackage{listings}
\usepackage{tikz}
\usetikzlibrary{fit, backgrounds, calc, positioning, arrows.meta, shapes.geometric, shadows}

\definecolor{vsc_key}{RGB}{4, 81, 165}      
\definecolor{vsc_string}{RGB}{163, 21, 21}   
\definecolor{vsc_punct}{RGB}{0, 0, 0}        
\definecolor{vsc_number}{RGB}{9, 134, 88}    
\definecolor{vsc_bg}{RGB}{255, 255, 255}
\definecolor{vsc_yellow}{RGB}{255, 215, 0}   
\definecolor{vsc_pink}{RGB}{218, 112, 214}

\newcommand{\btS}[1]{\textcolor{orange!80!black}{\textbf{#1}}} 
\newcommand{\btA}[1]{\textcolor{gray!70}{#1}}

\lstdefinelanguage{json}{
    basicstyle=\scriptsize\ttfamily,  
    showstringspaces=false,
    breaklines=true,
    frame=single,                          
    backgroundcolor=\color{vsc_bg},
    stringstyle=\color{json_string},
    keywords={type, children, name, args},
    keywordstyle=\color{vsc_string}\bfseries,
    numbers=none,                         
    literate=
     *{:}{{{\color{vsc_punct}{:}}}}{1}
      {,}{{{\color{vsc_punct}{,}}}}{1}
      {\{}{{{\color{vsc_key}{\{}}}}{1}
      {\}}{{{\color{vsc_key}{\}}}}}{1}
      {[}{{{\color{vsc_key}{[}}}}{1}
      {]}{{{\color{vsc_key}{]}}}}{1}
      {"}{{{\color{vsc_string}{"}}}}{1},
}

\begin{document}

\title{
Learning Structured Robot Policies from Vision-Language Models via Synthetic Neuro-Symbolic Supervision
%Automatically Generated Training Data for End-to-End Synthesis of High-Level Task Planner
}
%\author{Anonymous Authors}
\author{Alessandro Adami$^{*,\mathsection}$, Tommaso Tubaldo$^{\dagger}$, Marco Todescato$^\dagger$, Ruggero Carli$^*$, Pietro Falco$^*$
%<-this % stops a space
\thanks{Project co-funded by the European Union – Next Generation Eu - under the National Recovery and Resilience Plan (NRRP), Mission 4 Component 2, Investment 3.3 – Decree no. 630 (24th April 2024)  of Italian Ministry of University and Research; Concession Decree no. 1956 of 05th December 2024 adopted by the Italian Ministry of University and Research, within the national PhD Program in Autonomous Systems (XL cycle)}
 %<-this % stops a space
\thanks{\\$*$  University of Padova, Dept. of Information Engineering, Italy.\\
$\dagger$ Fraunhofer Italia Research, 39100, Bozen, Italy.\\
$\mathsection$ Polytechnic of Bari Dept. of Electrical and Information Engineering, Italy.}}% <-this % stops a space
%\thanks{Manuscript received April 19, 2021; revised August 16, 2021.}}

% The paper headers
%\markboth{Journal of \LaTeX\ Class Files,~Vol.~14, No.~8, August~2021}%
%{Shell \MakeLowercase{\textit{et al.}}: A Sample Article Using IEEEtran.cls for IEEE Journals}

%\IEEEpubid{0000--0000/00\$00.00~\copyright~2021 IEEE}
% Remember, if you use this you must call \IEEEpubidadjcol in the second
% column for its text to clear the IEEEpubid mark.

\maketitle

\begin{abstract}
Vision-Language Models (VLMs) have recently demonstrated strong capabilities in mapping multimodal observations to robot behaviors. However, most current approaches rely on end-to-end visuomotor policies that remain opaque and difficult to analyze, limiting their use in real-world robotic applications. In contrast, classical robotic systems often rely on structured policy representations that provide interpretability, modularity, and reactive execution.
This work investigates how foundation models can be specialized to generate structured robot policies grounded in multimodal perception, bridging high-dimensional learning and symbolic control. We propose a neuro-symbolic approach in which a VLM synthesizes executable Behavior Tree policies from visual observations, natural language instructions, and structured system specifications. To enable scalable supervision without manual annotation, we introduce an automated pipeline that generates a synthetic multimodal dataset of domain-randomized scenes paired with instruction–policy examples produced by a foundation model.
By decoupling structured task decomposition under constrained symbolic grammars from hardware-specific motor control, we demonstrate that a 12B-parameter model can learn structured spatial-symbolic mappings required for executable BT synthesis, solely through in-silico supervision. Real-world physical experiments on two heterogeneous robotic manipulators confirm that these structurally constrained policies achieve zero-shot transfer to real-world environments. The results emphasize that the data bottleneck in robotic planning can be bypassed by procedurally synthesizing high-fidelity, neuro-symbolic training data.
%Behavior Trees (BTs) offer a modular and reactive representation for robotic task execution, yet their direct synthesis from multimodal real-world observations remains challenging. We present an end-to-end framework for generating BTs from RGB synthetic observations, natural-language instructions, and structured system specifications by fine-tuning an open-source vision-language model for constrained symbolic output. To supervise this process, we define an automated pipeline for the generation of a fully synthetic dataset of domain-randomized tabletop scenes paired with symbolic metadata and automatically generated instruction--BT annotations, leveraging zero real-world training samples. The framework enforces structural priors on the target trees, including atomic action decomposition, reactive guarding, and spatial offsetting, to improve validity and execution robustness. We instantiate the approach by fine-tuning Pixtral-12B vision-language model, and validate it through ablation studies and real-world experiments. Results on a Franka Emika Panda and a UR5e demonstrate that our framework achieves task success on in-distribution commands, proving that high-level robotic logic can be successfully transferred from purely synthetic supervision to physical hardware with zero real-world training samples.
\end{abstract}

\begin{IEEEkeywords}
Vision-Language Models (VLMs), Behavior Trees, Robot Learning, Synthetic-to-Real Transfer, Task Planning, Symbolic Policies. %Reactive Control, Fine-Tuning
\end{IEEEkeywords}

\section{Introduction}
The deployment of general-purpose robots in robotics environments requires the seamless integration of multimodal perception and high-level decision-making. While recent advancements in Vision-Language Models (VLMs) and Vision-Language-Action (VLA) models~\cite{brohan2023rt2visionlanguageactionmodelstransfer, kim2024openvlaopensourcevisionlanguageactionmodel,huang2024a3vlmactionablearticulationawarevision, driess2023palmeembodiedmultimodallanguage,black2026pi0visionlanguageactionflowmodel} have shown remarkable capabilities in mapping observations to actions, they are fundamentally bottlenecked by the \textbf{data scarcity problem}. Collecting high-quality, human-annotated robotic demonstrations in the real world is expensive, non-scalable, and often lacks the logical diversity required for robust task planning.

In response to this challenge, this work investigates the feasibility of \textbf{learning structured robot policies entirely from synthetic neuro-symbolic supervision}. Rather than relying on labor-intensive real-world datasets, we propose a methodology that specializes a medium-scale VLM to produce interpretable, reactive, and executable robot policies without real-world training samples. In this framework, Behavior Trees (BTs)~\cite{IOVINO2022104096, ColledanchiseLTL} are utilized not as a static planning tool, but as the symbolic target representation that provides the modularity and logical transparency missing from end-to-end "black-box" motor policies~\cite{wake2025vlmdrivenbehaviortreecontextaware, adami2026real2simbasedactiveperception}.

\begin{figure*}[!t]
    \centering
    \includegraphics[width=\textwidth]{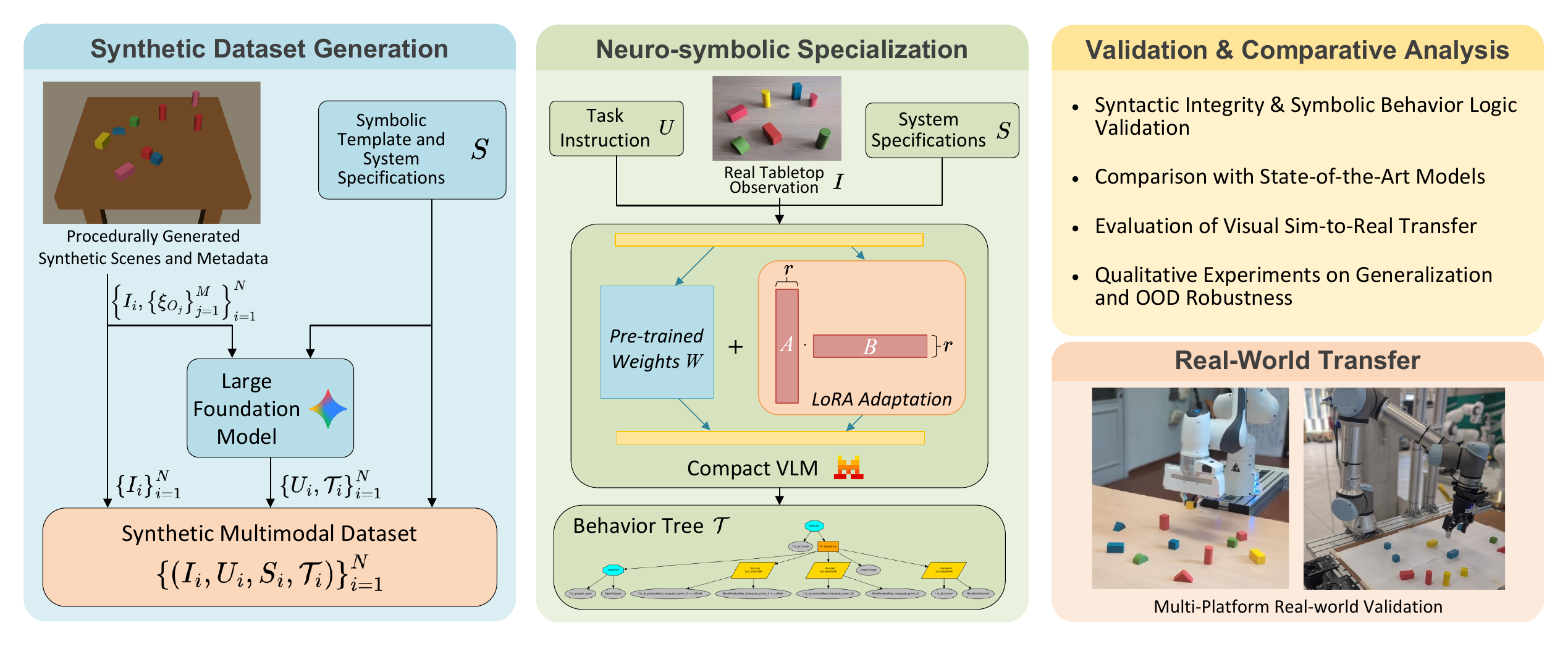}
    \caption{\textbf{Overview of the proposed framework.} First, high-fidelity simulation is used to procedurally generate domain-randomized synthetic scenes and structured metadata. A large foundation model then converts these observations into a synthetic multimodal supervision dataset of task instructions and executable Behavior Trees. This dataset is subsequently used to fine-tune a 12B-parameter vision-language model for constrained symbolic policy synthesis. At inference time, the fine-tuned model receives a real-world observation, a task instruction, and system specifications, and outputs a Behavior Tree representing an interpretable and reactive robot policy for task execution.}
    \label{fig:main_image}
\end{figure*}

The core of our approach is an automated, two-stage data synthesis and specialization pipeline (Fig.~\ref{fig:main_image}). First, we leverage high-fidelity physics engines paired with large-scale foundation models to \textbf{procedurally generate 10,000 unique multimodal training samples} from scratch. This dataset bridges the visual-symbolic gap by pairing domain-randomized synthetic observations with automatically synthesized, logically consistent instruction--policy pairs. Second, we utilize this synthetic supervision to fine-tune an open-source 12B-parameter VLM, internalizing the structural priors and spatial arithmetic necessary for physical interaction.

Our results demonstrate that the logical grounding and spatial reasoning required for robotic task execution can be distilled from purely synthetic data. We show that a model trained exclusively in simulation can perform zero-shot logic transfer to real-world scenarios within the targeted manipulation domain, handling novel industrial components and referential ambiguity on multiple physical manipulators. 

The primary contributions of this work are focused on the \textbf{data-centric and training methodology} for multimodal robot learning:
\begin{itemize}
    \item \textbf{Scalable Synthetic Supervision:} An automated pipeline for generating large-scale multimodal robotic datasets, eliminating the need for real-world collection while ensuring logical and syntactic consistency between vision, language, and policy. This significantly reduces the human effort required to collect data for training.
    \item \textbf{Neuro-Symbolic Specialization:} A methodology for fine-tuning medium-sized VLMs through LoRA adaptation to output structured symbolic policies, internalizing architectural constraints such as reactive guarding and symbolic spatial offsetting as safety constraints.
    \item \textbf{Zero-Real-World-Data Validation:} Extensive physical experiments demonstrating that models trained with zero real-world samples achieve performance parity with human-prompted state-of-the-art models in structured task planning.
\end{itemize}

We validate the approach through training ablations and real-world experiments on two different manipulators, namely a Franka Emika Panda and a UR5e, showing that the synthesized BTs remain logically consistent across platforms and transfer from synthetic supervision to physical execution. This work assumes a modular perception-planning architecture. We delegate low-level 6D pose estimation to established external perception pipelines, allowing our fine-tuned VLM to focus entirely on high-level spatial reasoning and symbolic policy synthesis. The objective of this work is not end-to-end embodied perception, but the study of multimodal symbolic policy synthesis under structured scene grounding.
%The remainder of this paper is organized as follows. Section~\ref{Sec: related-works} provides an overview of the state-of-the-art in LLM-based task planning and Sim-to-Real transfer. Section~\ref{Sec: problem and solution} formalizes the problem statement and details the proposed Behavior Tree synthesis framework. Section~\ref{sec:dataset_creation} describes the automated pipeline for fully synthetic dataset generation. Section~\ref{Sec: Training} outlines the vision-language model architecture and the fine-tuning methodology. Section~\ref{Sec: Ablation} presents a comprehensive evaluation of the synthesized trees through systematic ablation studies. Section \ref{Sec: real-world} validates the framework on two physical robotic platforms. Finally, Section VIII concludes the paper and discusses future research directions.
The remainder of this paper is organized as follows. Section~\ref{Sec: related-works} provides an overview of the state-of-the-art in LLM-based task planning and Sim-to-Real transfer. Section~\ref{Sec: problem and solution} formalizes the problem statement and the proposed neuro-symbolic framework. Section~\ref{sec:dataset_creation} details the automated pipeline for large-scale synthetic dataset generation, while Section~\ref{Sec: Training} outlines the vision-language model architecture and the supervised fine-tuning methodology. Section~\ref{Sec: Ablation} presents a comprehensive evaluation of the synthesized trees through systematic ablation studies. Section~\ref{Sec: real-world} validates the framework on two physical robotic platforms. Section~\ref{Sec: Generalization} investigates the invariance and robustness of the model across novel objects and hardware, and Section~\ref{Sec: Limitations} provides a critical boundary analysis of the current failure modes. Finally, Section~\ref{Sec: conclusion} concludes the paper and discusses future research directions.

\textbf{To support reproducibility, the source code, synthetic dataset, and fine-tuned model weights will be made publicly available upon publication.}
%\textbf{In accordance with open science principles, all source code, the synthetic dataset, and the fine-tuned model weights will be made publicly available upon publication.}

\section{Related Works}\label{Sec: related-works}
    \subsection{Large Language Models for Behavior Tree Generation}
    Behavior Trees (BTs)~\cite{IOVINO2022104096, unifiedBTframework} have become a standard in robotic control~\cite{ColledanchiseLTL} due to their modularity, readability, and reactive nature. While traditionally manually designed by experts, recent research has shifted toward leveraging Large Language Models (LLMs) to automate BT generation~\cite{BTExpansionwithLLMs, CodeBT} and domains beyond natural language processing~\cite{pmlr-v205-huang23c}. Early studies often used LLMs to produce preliminary task information or to fill fixed-structure BT templates. More recent approaches, such as~\cite{LLMBrain}, directly generate full, executable BTs from text descriptions by fine-tuning models~\cite{alpaca}. Similarly,~\cite{izzo2026btgenbot2efficientbehaviortree} demonstrates that lightweight LLMs (under 7B parameters) can generate effective robot behaviors when fine-tuned on high-quality datasets curated from existing robotic projects. Frameworks like~\cite{LLMasBTPlanner} further extend these capabilities to complex assembly tasks, using in-context learning to decompose instructions into subgoal sequences before generating parameterized task plans. While earlier works~\cite{LLMBrain} focused on the linguistic mapping of tasks to trees, they often lacked visual grounding, relying instead on pre-parsed object lists. Our work bridges this gap by moving from text-to-BT to vision-to-BT, eliminating the need for a separate symbolic perception pipeline.

    % \subsection{Vision-Language-Action Models and API calling}\label{Sec: VLA}
    % Vision-Language-Action (VLA) models have evolved from generating robotic API calls \cite{liangcodeaspolicies2023} to reasoning about physical space within embodied multimodal architectures \cite{driess2023palmeembodiedmultimodallanguage}. Generalist policies such as RT-2 \cite{brohan2023rt2visionlanguageactionmodelstransfer} and the open-source OpenVLA \cite{kim2024openvlaopensourcevisionlanguageactionmodel} utilize co-fine-tuning on large-scale trajectory data to map visual observations directly to tokenized actions. However, while these end-to-end systems excel at direct control, they often function as "black boxes" and lack the inherent reactivity and logical transparency of hierarchical structures like Behavior Trees. To bridge this gap, some approaches introduce object-centric awareness to translate VLM reasoning into robot-agnostic action primitives \cite{huang2024a3vlmactionablearticulationawarevision}. Our work addresses a significant research gap by fine-tuning medium-sized open-source models specifically for structured symbolic output rather than natural language or low-level motor commands.

    \subsection{Context-Aware Planning and Interactive Systems}
    The static nature of early LLM-generated plans often fails in complex environments, when their description is not accurate enough. To address this, recent frameworks incorporate Vision-Language Models (VLMs) directly into the execution loop. A notable advancement is the use of self-prompted visual conditions~\cite{wake2025vlmdrivenbehaviortreecontextaware}, where a VLM generates BT condition nodes expressed as free-form text. These conditions are then evaluated against egocentric images during execution, allowing the robot to branch its behavior based on visual feedback (e.g., checking if a cup is full before moving it). Following \cite{izzo2026btgenbot2efficientbehaviortree}, \cite{battistini2026multimodal} investigates fine-tuning compact VLMs for executable BT generation from RGB observations and natural-language instructions. Their supervision is derived through a teacher-driven pipeline built on a pre-existing real-world robotic dataset, and evaluation is performed in an embodied simulation environment. While this extends BT generation to the multimodal setting, it still relies on recorded real-world robotic data. In contrast, our work demonstrates that such supervision can be achieved with zero real-world samples, relying instead on high-fidelity synthetic scene generation. 
    Recent work has extended VLM-generated BTs from task execution to active perception. \cite{adami2026real2simbasedactiveperception} proposed a Real2Sim framework in which a VLM synthesizes BTs from atomic interaction primitives to perform physical interactions in order to estimate physical parameters such as mass and surface friction, demonstrating that atomic BT primitives are sufficient for complex, contact-rich exploration tasks. More broadly, it reflects a shift from conventional Sim-to-Real transfer, centered on visual appearance adaptation, to \emph{logical grounding} across domains. Under this perspective, the key challenge is not only object recognition, but also the correct mapping of visual-spatial relations into symbolic BT parameters. Our approach addresses this problem through high-fidelity synthetic data designed to supervise such logical grounding directly.

    \subsection{Sim-to-Real Transfer in Vision-Language Models}
    Bridging the Sim2Real gap remains a central challenge for deploying VLMs in robotic manipulation. Recent work has shown that carefully designed synthetic data can improve spatial reasoning and enable transfer to real-world tasks without costly manual annotation \cite{rizzoli2026syntheticscenesrealperformance}. Complementary efforts address the visual domain gap through neuro-symbolic image adaptation, where structured knowledge is used to preserve semantic spatial relations during sim-to-real translation \cite{youssef2026ontologyguideddiffusionzeroshotvisual}. In parallel, architectures such as \cite{zhao2026simreal} combine high-level planning with low-level validation to prioritize task-relevant dynamics and support zero-shot transfer of manipulation skills, and \cite{peng2025roboseekneedinteractobjects} relies on continuous real-world interaction with the objects. Our work follows this direction, but focuses on synthetic supervision for the generation of reactive, verifiable Behavior Trees rather than motor policies or visual adaptation.

    \subsection{End-to-End VLA Models vs. Symbolic Task Planning}
    Recent trends in robot learning have seen the rise of Vision-Language-Action (VLA) models, such as~\cite{kim2024openvlaopensourcevisionlanguageactionmodel} and~\cite{black2026pi0visionlanguageactionflowmodel}, which utilize large-scale imitation learning to map multimodal observations directly to continuous motor trajectories. While these end-to-end approaches demonstrate impressive motor fluency in short-horizon tasks, they function as high-dimensional black boxes. They lack an explicit representation of task logic, making it difficult to verify safety constraints or implement reactive recovery behaviors (e.g., 'retry if grasp fails').

    Furthermore, because VLA models typically output low-level action chunks or trajectories rather than discrete logical steps, a direct empirical comparison between our framework and end-to-end policies is structurally incompatible. Our approach operates at the symbolic orchestration layer, synthesizing a verifiable and interpretable Behavior Tree that delegates motor execution to a primitive library. This decoupling allows for formal logic checks, such as the mandatory z-axis approach offsets enforced in our pipeline, which are not natively guaranteed by trajectory-prediction models.

    \subsection{Data Synthesis and Verification Strategies}
    The effectiveness of LLM-based planners strongly depends on training data quality and scale. Recent work has shown that synthetic data can improve the learning of task logic \cite{li2024studytrainingdevelopinglarge}, while large-scale synthetic supervision has also proven effective for robust sim-to-real transfer in perception tasks such as 6D pose estimation and semantic grasping \cite{foundationposewen2024,DeepObjPoseEstimation}. We build on this line of work by using synthetic data not only for geometric supervision, but also for high-level policy synthesis in the form of BTs. To improve deployment reliability, prior studies have additionally introduced multi-stage verification pipelines combining syntactic checks with simulation-based validation before execution on real hardware \cite{izzo2026btgenbot2efficientbehaviortree}.

    \subsection{Structured Prompting and Symbolic Reasoning}
    Structured specifications have been shown to improve Large
    Multimodal Models (LMMs) performance in domain-specific planning tasks relative to unconstrained natural-language prompting \cite{hu2024chainofsymbolpromptingelicitsplanning}. Following this principle, we adopt a JSON-based specification that encodes the primitive library, safety constraints, and dynamic scene metadata. In contrast to text-only in-context planning approaches \cite{LLMasBTPlanner}, our formulation extends structured prompting to the multimodal setting and requires the model to perform symbolic operations, such as approach pose offsetting, directly within the generated Behavior Tree. This design links structured task decomposition under constrained symbolic grammars to the geometric constraints of robot manipulation.

    In contrast to prior works that primarily focus on end-to-end policy learning or template-based planning, this work investigates how multimodal foundation models can be specialized to generate structured symbolic robot policies that preserve interpretability and reactivity. Our approach differs from existing multimodal BT generation methods by relying exclusively on synthetic supervision, eliminating the need for real-world demonstrations while maintaining logical grounding and executable policy structure.

\section{Problem and Proposed Solution}\label{Sec: problem and solution}

\subsection{Problem Statement: The Grounding and Reliability Gap}\label{sec:problem_statement}
    Despite recent progress in multimodal BT generation, existing methods remain limited by incomplete symbolic grounding, dependence on real-world robotic data, or evaluation restricted to simulation~\cite{LLMBrain, izzo2026btgenbot2efficientbehaviortree}. In practice, generating executable BTs from visual observations and language requires more than semantic recognition: the model must map visual-spatial relations into valid symbolic structures that satisfy geometric and architectural constraints of robotic execution. This unresolved grounding and reliability gap leads to the following bottlenecks in robotics environments:
    
    \begin{enumerate}
        \item \textbf{Decoupled Perception and Planning ($P_1$):} Traditional systems separate visual feature extraction from logical reasoning, leading to error propagation and high latency.
        \item \textbf{Heuristic Inconsistency ($P_2$):} General-purpose models frequently struggle to adhere to strict robotic protocols, such as mandatory $z$-offsets for collision-free trajectories.
        \item \textbf{Execution Fragility ($P_3$):} Linear action sequences lack the inherent reactivity of Behavior Trees (BTs), preventing autonomous recovery from environmental perturbations.
        \item \textbf{Computational Opacity ($P_4$):} Reliance on proprietary, high-parameter models limits reproducibility and real-time deployment on robotic edge devices.
    \end{enumerate}

\subsection{Problem Formulation and Symbolic Grounding}\label{sec:sys_assumptions}
    Let $\mathcal{I}$ be the space of high-dimensional RGB observations and $\mathcal{U}$ the space of natural language instructions. Furthermore, let $\mathcal{S}$ be the set of structured system specifications, where each $S \in \mathcal{S}$ encapsulates the available primitive library, safety constraints, and dynamic object metadata.
    Our framework defines the synthesis task as a mapping function $f_\theta: \mathcal{I} \times \mathcal{U} \times \mathcal{S} \to \Pi$, where $\Pi$ is the set of all valid Behavior Trees. 
    A generated BT $\mathcal{T} \in \Pi$ is considered \textit{structurally sound} if it satisfies the formal grammar of the JSON-BT schema and \textit{behaviorally grounded} if the computed parameters in $\mathcal{T}$ align with the spatial distribution of objects in $I\in\mathcal{I}$.
    
    To isolate the challenges of logical synthesis and spatial reasoning, we assume the existence of a high-level scene graph provided as symbolic metadata. Specifically, each object $O_i$ in the visual scene is associated with a dynamic pose label $\xi_{O_i}$ within the system specification $S$. Unlike traditional static planners, these poses are \textit{dynamic variables}: if an object is displaced during execution, the metadata is updated, and the synthesized BT remains valid by referencing the symbolic label rather than a hard-coded coordinate. This formulation allows the VLM to focus on the mapping of human instructions to geometric relationships, assuming low-level localization is handled by an external perception backbone.

\subsection{Behavior Trees}
Behavior Trees (BTs) are a modular control formalism for organizing task execution in robotics and AI~\cite{ColledanchiseLTL,IOVINO2022104096}. Formally, a BT is a directed rooted tree in which the internal nodes are control-flow nodes and the leaf nodes are execution nodes. Using standard tree terminology, each node except the root has exactly one parent, while control-flow nodes have one or more children.

Execution in a BT starts from the root, which periodically sends \emph{ticks} to its children at a given frequency. A node is executed if and only if it receives a tick, and in response it returns one of three statuses to its parent: \texttt{Running}, if execution is still under way; \texttt{Success}, if the objective has been achieved; or \texttt{Failure}, otherwise. The behavior of the full tree is therefore determined by how ticks are propagated through the hierarchy and how return statuses are combined by the control-flow nodes.

In the classical formulation, BTs are defined by three principal classes of control-flow nodes, namely \texttt{Sequence}, \texttt{Selector}, and \texttt{Parallel}, and two classes of execution nodes, namely \texttt{Action} and \texttt{Condition}. A \texttt{Sequence} node ticks its children from left to right and returns \texttt{Failure} or \texttt{Running} as soon as one child returns that status; it returns \texttt{Success} only if all children return \texttt{Success}. Dually, a \texttt{Selector} node ticks its children from left to right and returns \texttt{Success} or \texttt{Running} as soon as one child returns that status; it returns \texttt{Failure} only if all children fail.

A \texttt{Parallel} node ticks multiple children concurrently and evaluates their returned statuses according to a user-defined success condition, typically expressed as the minimum number of children that must succeed. Finally, \texttt{Action} nodes implement commands that affect the environment, whereas \texttt{Condition} nodes test properties of the current state and typically return immediately with either \texttt{Success} or \texttt{Failure}.

\subsection{Proposed Solution: End-to-End Behavior Tree Synthesis}
    To address the limitations identified in Section~\ref{sec:problem_statement}, we propose a framework in which the \textbf{Pixtral-12B}~\cite{agrawal2024pixtral12b} Vision-Language Model is fine-tuned to perform direct synthesis of Behavior Trees from multimodal inputs ($P_1$, $P_4$).

    Our solution treats BT generation as a \textbf{constrained multimodal optimization task}. By synthesizing a structured JSON-encoded tree rather than raw motor commands, we provide a \textbf{symbolic execution trace}. This allows for pre-execution syntax-level verification, where a static analyzer can guarantee that the generated policy respects safety invariants ($P_2$) before execution. 

    A key element of the proposed framework is the \textbf{fully synthetic construction of the fine-tuning dataset} (Sec.~\ref{sec:dataset_creation}), which enables explicit control over scene composition and ensures annotation fidelity. Compared with labor-intensive real-world data collection, the automatic generation of synthetic observations and paired natural language instruction-BT samples provides a scalable and systematic annotation pipeline with minimal manual intervention.
    
    We enforce three core structural paradigms through synthetic dataset creation:
    \begin{itemize}
        \item \textbf{Reactive Guarding:} Interaction nodes (e.g., \texttt{CloseGripper}) are nested within a \texttt{Selector} or \texttt{Parallel} with \texttt{Condition} checks (e.g., \texttt{is\_at\_pose}) ($P_3$).
        \item \textbf{Spatial Offsetting:} The model is trained to compute derived poses by applying a positive $z$-offset to base object coordinates for approach and lift phases ($P_2$).
        \item \textbf{Structural Homogeneity:} To prevent logic depth errors ($P_2$), we enforce a flat hierarchy where composite nodes of the same type are strictly prohibited.
    \end{itemize}

\subsection{Atomic Primitive Decomposition}\label{Sec: Atomic Actions}
    We decompose complex tasks into \textit{atomic primitives} as did in~\cite{adami2026real2simbasedactiveperception}, in contrast with high-level actions of works described in Sec.~\ref{Sec: related-works}, providing three distinct advantages:
    \begin{itemize}
        \item \textbf{Explainability:} Every step of the robot's reasoning is traceable through the BT leaf nodes.
        \item \textbf{Execution Robustness:} Elementary checks allow for high-frequency re-evaluation, triggering recovery branches ($P_3$) upon failure.
        \item \textbf{Model Efficiency:} Reducing the output space to a finite set of primitives lowers generation complexity, allowing the 12B-parameter VLM to compete with larger closed-source models ($P_4$, Sec.~\ref{sec: SOTA}).
    \end{itemize}
    Tab.~\ref{tab:primitives} defines the set of atomic actions and condition checks used in our framework.
    
    \begin{table}[]
    \centering
    \begin{tabular}{lll}
    \hline
    \textbf{Primitive} & \textbf{Type} & \textbf{Functionality Description} \\ \hline
    \texttt{MovePose} & Action & Moves end-effector to a 6D goal pose. \\
    \texttt{OpenGripper} & Action & Fully opens the parallel gripper fingers. \\
    \texttt{CloseGripper} & Action & Closes the gripper until contact or limit. \\
    \texttt{MoveDown} & Action & Moves vertically down. \\
    \texttt{is\_at\_pose} & Condition & Verifies if the gripper is in a pose. \\
    \texttt{is\_grasped} & Condition & Checks if an object is in the gripper. \\
    \texttt{is\_contact} & Condition & Binary check for force sensor feedback. \\
    \texttt{is\_at\_home} & Condition & Confirms if the robot is in home state. \\ \hline
    \end{tabular}
    \vspace{4pt}
    \caption{Library of Atomic Robotic Primitives.}
    \label{tab:primitives}
    \end{table}

\subsection{Formal Structural Constraints}
    To address the requirements of safety and reactivity ($P_2, P_3$), we define a formal grammar for the synthesized tree $\mathcal{T}$. Let internal nodes $N_{comp} \in \{\text{Sequence, Selector, Parallel}\}$ and leaf nodes $L$ consist of Actions $A$ and Conditions $C$.

\subsubsection{\textbf{Formal JSON-BT Grammar}}
To enable static syntax verification before robotic execution, we restrict the VLM output to a strictly defined grammar. Let $\mathcal{T}$ denote a valid Behavior Tree subtree:
\begin{equation}
\begin{aligned}
    \mathcal{T} &::= \text{Composite}(c_1, \dots, c_k) \mid \text{Leaf} \\
    \text{Composite} &::= \{ \text{\texttt{"type"}}: C, \text{\texttt{"children"}}: [\mathcal{T}_1, \dots, \mathcal{T}_k] \} \\
    \text{Leaf} &::= \{ \text{\texttt{"type"}}: L, \text{\texttt{"name"}}: \gamma, \text{\texttt{"args"}}: [\dots] \}
\end{aligned}
\end{equation}
where $C \in \{\texttt{Sequence}, \texttt{Selector}, \texttt{Parallel}\}$, $L \in \{\texttt{Action}, \texttt{Condition}\}$, and $\gamma$ belongs to the atomic primitive library. This strict formalization guarantees that any JSON successfully parsed into this grammar will not trigger runtime parsing errors on the physical robot.

\subsubsection{\textbf{Reactivity via Parallel and Selector Verification}}
    Unlike linear planners, our model encapsulates motion actions within a \texttt{Parallel} and \texttt{Selector} composite nodes, with a corresponding condition. 
    For example, for a goal pose $\xi$, the motion subtree $\mathcal{T}_{move}$ is defined as:
    \begin{equation}
\begin{split}
\mathcal{T}_{move} = \text{Parallel}(
\text{Condition}(\text{is\_at\_pose}, \xi), \\
\quad \text{Action}(\text{MovePose}, \xi)
)
\end{split}
\end{equation}
    This ensures motion and perceptual verification are evaluated concurrently, allowing the BT to succeed immediately upon meeting the physical threshold.
    
\subsubsection{\textbf{Symbolic Arithmetic}}
    To resolve $P_2$, the model performs symbolic arithmetic within the JSON schema. For an object $O$ at pose $\xi_O$, the model must synthesize an \textit{approach pose} $\xi_a$:
    \begin{equation}
        \xi_a = \mathbf{T}_{off} \cdot \xi_O = 
        \begin{bmatrix} 
            \mathbf{R} & \mathbf{t} + \Delta\mathbf{z} \\ 
            \mathbf{0}^T & 1 
        \end{bmatrix}
    \end{equation}
    where $\Delta\mathbf{z} = [0, 0, z_{off}]^T \text{ with } z_{off} > 0$ and where $\mathbf{R}$ and $\mathbf{t}$ respectively define the rotation and the position of the object.
    By grounding the BT in symbolic labels, the framework ensures that reactive checks remain valid under environmental perturbations, as the check node re-evaluates the dynamic label $\xi_O$ at runtime.
    
\subsubsection{\textbf{The Flat Hierarchy Constraint}}
    To prevent the model from hallucinating infinite recursive loops, a common failure mode in LLM-based planners, we enforce a strict structural homogeneity constraint during synthetic supervision. For any composite parent node $n_p \in \mathcal{T}$ and its direct children $c_i$, the following invariant must hold:
\begin{equation}
    \text{type}(n_p) \neq \text{type}(c_i) \quad \forall c_i \in \text{children}(n_p)
\end{equation}
This mathematically prohibits nested redundancies (e.g., a \texttt{Sequence} cannot be a child of another \texttt{Sequence}). By penalizing deep, redundant nesting, the model is forced to prioritize flat, wide hierarchies that maximize execution interpretability ($P_1$).

\section{Synthetic Dataset Generation}\label{sec:dataset_creation}
The efficiency and scalability of our framework stem from an automated pipeline designed to \textbf{generate a fully synthetic dataset} for Supervised Fine-Tuning (SFT), \textbf{eliminating the requirement for real-world data collection and labor-intensive manual annotation}. This two-stage process integrates high-fidelity simulation with large-scale multimodal reasoning to produce a diverse supervision signal. First, we utilize the MuJoCo physics engine to procedurally generate 10,000 domain-randomized tabletop scenes, capturing the spatial variance necessary for robust visual grounding. Second, a high-capacity foundation model is employed to interpret these scenes and synthesize logically consistent natural-language instructions paired with executable Behavior Trees. By strictly controlling the scene composition and the resulting symbolic logic, we provide a structured dataset that forces the model to internalize the spatial arithmetic and architectural priors required for reactive robotic execution, all while maintaining a purely in-silico training regime.
\subsection{Visual Scenes Generation}
\begin{figure}[]
    \centering
    \includegraphics[width=0.24\linewidth]{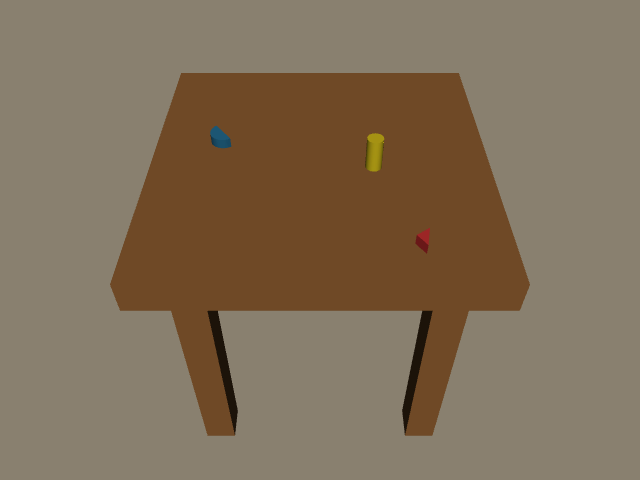}
    \includegraphics[width=0.24\linewidth]{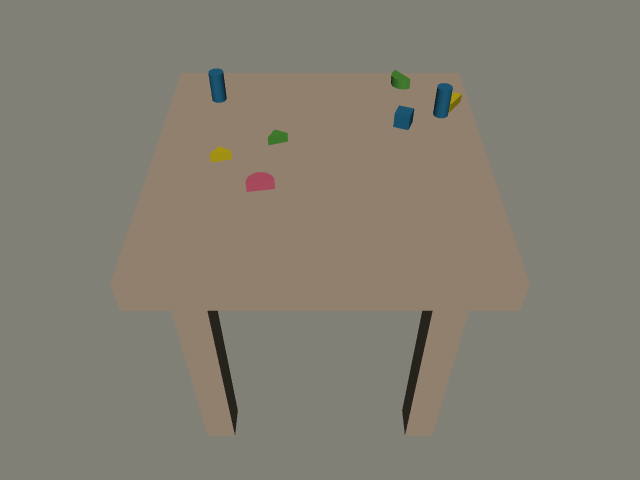}
    \includegraphics[width=0.24\linewidth]{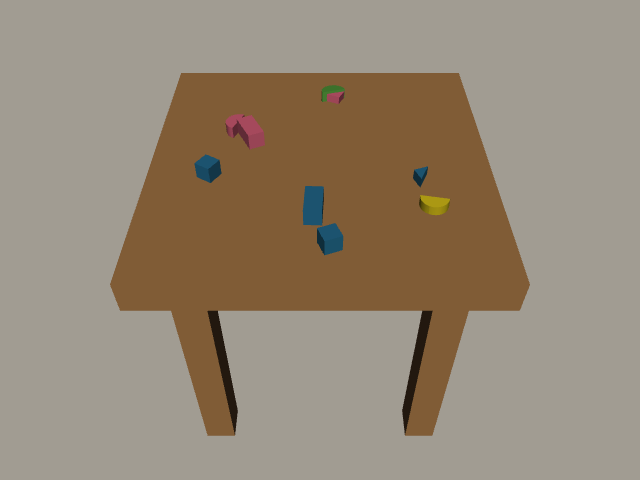}
    \includegraphics[width=0.24\linewidth]{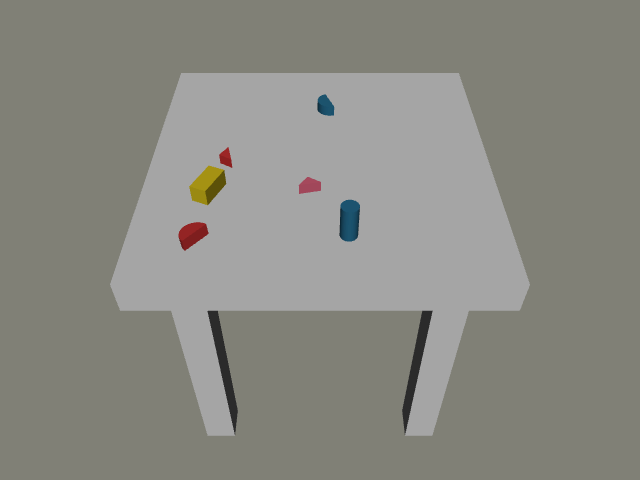}
    \includegraphics[width=0.24\linewidth]{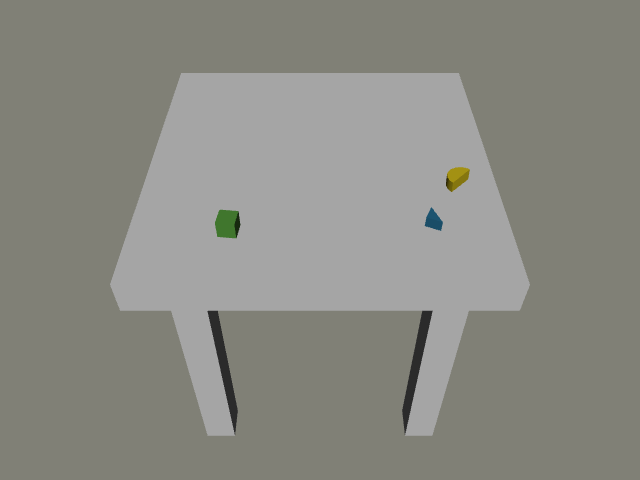}
    \includegraphics[width=0.24\linewidth]{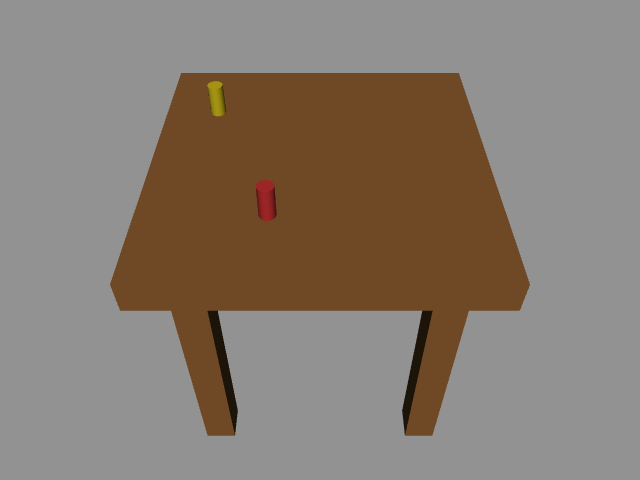}
    \includegraphics[width=0.24\linewidth]{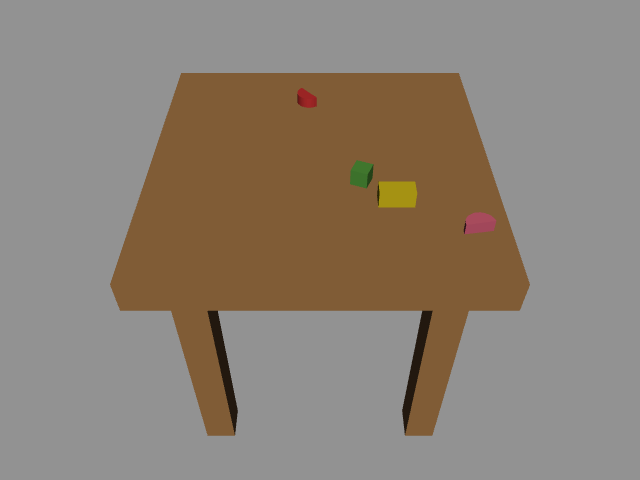}
    \includegraphics[width=0.24\linewidth]{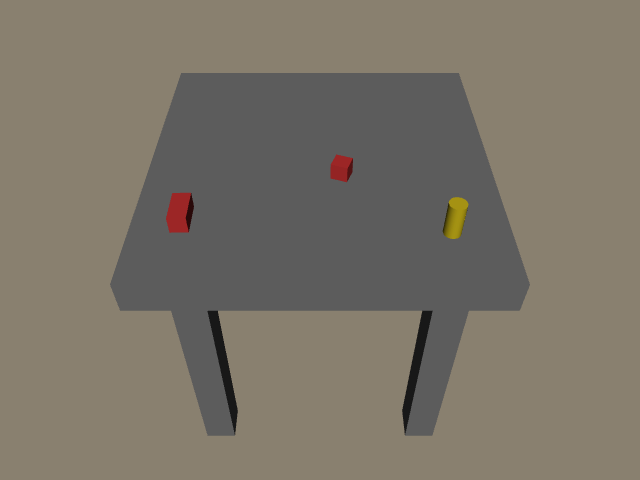}
    \caption{Examples of synthetic tabletop scenes used in dataset generation.}
    \label{fig: synthetic images}
\end{figure}

\textbf{The visual components of the dataset were generated in simulation} using the MuJoCo physics engine \cite{todorov2012mujoco}. To bridge the visual sim-to-real gap, we implemented Domain Randomization during dataset generation~\cite{tobin2017domainrandomizationtransferringdeep}. \textbf{We procedurally generated 10,000 unique tabletop scenes} (Fig.~\ref{fig: synthetic images}) featuring various geometric entities (e.g., prisms, parallelepipeds, cylinders) with randomized colors and spatial configurations.

As stated in Sec.~\ref{sec:sys_assumptions}, the poses of the scene objects $\left\{\xi_{O_j}\right\}_{j=1}^{M}$ are extracted for each scene and encoded as symbolic metadata used during training. In addition to object poses, the metadata includes auxiliary workspace poses, denoted \texttt{temp\_pose}, that can be referenced as intermediate waypoints during manipulation, for instance to temporarily place an object.

The use of synthetic images is a deliberate methodological choice, supported by recent advancements in automated robotic planning. In particular,~\cite{adami2026real2simbasedactiveperception} demonstrated that VLMs can effectively generate executable Behavior Trees when prompted on synthetic interaction data, which represents a reliable simplification of the real-world scenario.
By leveraging 10,000 high-fidelity synthetic samples, our model learns to associate visual object features with precise spatial logic, such as the requirement for $z$-offsets, without the noise and data-collection bottleneck of real-world environments. This approach ensures that the resulting model is grounded in physical reality.

\subsection{Multimodal Prompt Schema}
Our training input is structured to provide the compact VLM with a representation of the robotic scene, task specifications, and system constraints, thereby approximating the operational context of a real robotic system. As detailed in Tab.~\ref{tab:dataset elements}, the input is partitioned into three key blocks: the visual scene $I\in\mathcal{I}$, the user instruction $U\in\mathcal{U}$, and the system specifications $S\in\mathcal{S}$.
Beyond the input blocks, the schema defines the target output as a structured Behavior Tree $\mathcal{T} \in \Pi$. This tree is represented as a nested JSON object that encapsulates the reactive logic and symbolic parameters required for task execution. By including the target $\mathcal{T}$ in the schema definition, we formalize the mapping $f_\theta(I, U, S) \to \mathcal{T}$. An example of the expected BT-JSON structure for a picking task is provided in Fig.~\ref{List: BT JSON example}.
\begin{table}[]
\centering
\begin{tabular}{lp{5.5cm}}
\hline
\textbf{Element} & \textbf{Description and Role} \\ \hline
\textbf{Image ($I$)} & RGB rendering from MuJoCo, providing spatial context and object localization data (compacted as 224$\times$224 during training phase). \\
\textbf{Instruction ($U$)} & A human-centric command (e.g., "Place the pink prism on the green box") defining the high-level goal. \\
\textbf{Role \& Objective ($S$)} & Part of $\mathcal{S}$, defining the agent as an expert interpreter tasked with generating BTs. \\
\textbf{Node Library ($S$)} & A formal list of atomic Actions (\texttt{MovePose}, \texttt{OpenGripper}) and Conditions (\texttt{is\_grasped}) available to the agent. \\
\textbf{Metadata ($S$)} & A list of unique object identifiers (e.g., \texttt{pink\_prism\_1}) and home for the robot. \\
\textbf{Safety Rules ($S$)} & Explicit instructions for mandatory $z$-offsets and the enforcement of a flat hierarchy. \\ 
\textbf{Target BT ($\mathcal{T}$)} & The ground-truth JSON-encoded symbolic representation of the task logic, serving as the supervision signal for fine-tuning. \\ \hline
\end{tabular}
\vspace{4pt}
\caption{Elements of the Multimodal Training Dataset.}
\label{tab:dataset elements}
\end{table}

\subsection{Multimodal Prompt Generation}\label{Sec: multimodal prompt geenration}
The core of our synthetic supervision strategy is an automated annotation pipeline that transforms raw MuJoCo renders and symbolic metadata into high-quality training tuples $(I,U,S,\mathcal{T})$.
For each synthetic scene, we \textbf{automatically generate a natural-language instruction} and its \textbf{corresponding ground-truth Behavior Tree} by means of a large foundation model. Specifically, we employ \textbf{Gemini 3 Flash} \cite{team2023gemini} to produce 10,000 instruction-BT pairs through an iterative one-shot generation process, in which the scene-level information serves as the conditioning context for the generation of instruction $U$ and associated structured Behavior Tree $\mathcal{T}$.

\begin{figure}[t]
\begin{lstlisting}[language=json]
{
  "type": "Selector",
  "children": [
    { "type": "Condition", "name": "is_grasped" },
    {
      "type": "Sequence",
      "children": [
        { "type": "Action", "name": "MovePose", "args": 
            ["object_metadata"] 
        },
        { "type": "Action", "name": "CloseGripper" }
      ]
    }
  ]
}
\end{lstlisting}
\caption{Representation of the target Behavior Tree $\mathcal{T}$ in the prompt schema.}
\label{List: BT JSON example}
\end{figure}
At each iteration, the generation process is decomposed into two consecutive stages. In the first stage, the model generates the task instruction $U$ by conditioning on the scene-level information, namely the rendered image and the associated symbolic metadata. To encourage linguistic diversity and avoid overly templated commands, this step is performed with a high sampling temperature, i.e., $T=1.4$, ensuring a wide range of human-centric expressions. In the second stage, the model generates the corresponding Behavior Tree $\mathcal{T}$ using a low sampling temperature, i.e., $T=0.2$, thus favoring structural consistency and reducing variability in the symbolic output. This second-generation step is conditioned on the previously generated task instruction $U$, the same scene-level information, and a structured JSON schema that provides a one-shot reference example specifying both system specifications $S$, the organization of the multimodal training input, and the expected structure of the target BT.

% This automated annotation strategy ensures that the language instruction, the scene configuration, and the target BT remain semantically aligned. As a result, the dataset provides internally consistent supervision for learning symbolic grounding, namely the mapping from visual object relationships to executable parameters within the generated tree.

To quantify computational efficiency, we report average generation times split into simulation and annotation phases. Image acquisition, comprising MuJoCo environment initialization, RGB rendering with metadata collection, and process termination, requires $T_{\text{Img}} \approx 0.42$\,s per episode. Semantic annotation and BT generation via the Gemini API introduce an additional latency of $T_{\text{JSON}} \approx 10.87$\,s per sample, though this estimate reflects an empirical average due to inherent cloud latency variability.

Collecting 10,000 real-world robotic demonstrations typically requires a huge amount of human teleoperation and annotation. Our primary contribution is a procedural data engine that synthesizes 10,000 high-fidelity, neuro-symbolic training samples, effectively bypassing the robotics data bottleneck.

\subsection{Annotation Fidelity and Quality Validation}
To assess the fidelity of the automated annotation pipeline, we performed a two-stage quality validation over the full 10,000-sample dataset. First, an automated script verified the syntactic validity of all generated BTs against the JSON-BT grammar and cross-checked all symbolic arguments against the scene metadata. This process identified only two labeling inconsistencies across the entire dataset: one malformed object identifier (a misspelled \texttt{parallelelepiped} label) and one incorrect index suffix (\texttt{obj\_0} in place of \texttt{obj\_2} present in metadata). Both were isolated to single samples and did not propagate structurally. Second, semantic and logical correctness (specifically, whether the synthesized BT encodes the correct task logic for the paired instruction and scene) was validated through manual inspection of a stratified random batch of samples drawn across task types and scene complexities. No systematic logical errors were identified in this batch. These results indicate that the Gemini-based annotation pipeline produces highly consistent supervision, and that labeling noise in the training set is negligible.

\section{Model Training}\label{Sec: Training}
The training phase of our framework is designed to specialize the complex multimodal reasoning and structural priors contained within the synthetic dataset into a specialized controller. We formulate the learning process as a \textbf{Supervised Fine-Tuning (SFT)} task, where the objective is to optimize the mapping between the high-dimensional visual input $I$, the linguistic intent $U$, the system constraints $S$, and the target symbolic execution trace $\mathcal{T}$. By specializing a general-purpose vision-language model on our procedurally generated logic, we aim to bridge the grounding gap, moving from \textbf{generic natural-language} to deterministic, safety-critical robotic planning. To achieve this within the computational constraints of edge-deployable hardware, we utilize a \textbf{memory-efficient adaptation strategy} that preserves the foundational multimodal knowledge of the base model while injecting the specialized domain expertise required for Behavior Tree synthesis.

\subsection{Model Selection and Architecture}
    To address the challenges of computational opacity ($P_4$), we utilize \textbf{Pixtral-12B}~\cite{agrawal2024pixtral12b} as our foundational vision-language backbone. This open-source model integrates a 400M-parameter vision encoder with a 12B-parameter multimodal decoder. We employ \textbf{Low-Rank Adaptation (LoRA)}~\cite{hu2021loralowrankadaptationlarge} to specialize this general-purpose backbone for structured Behavior Tree synthesis. 

    To maximize the model's capacity for complex logical reasoning and symbolic arithmetic, we target a \textbf{heavy module configuration}. Unlike lite configurations that only adapt Attention layers, our implementation injects trainable low-rank matrices into the query, key, value, and output projections ($q, k, v, o$), alongside the Multi-Layer Perceptron (MLP) \texttt{gate\_proj}, \texttt{up\_proj}, and \texttt{down\_proj} modules. The model deployed for our real-world robotic experiments (Sec.~\ref{Sec: real-world}) was configured with a rank $r=16$, a scaling factor $\alpha=32$, and a dropout rate of $0.05$. This configuration effectively balances task-specific adaptation with the preservation of the base model's multimodal knowledge (Sec.~\ref{tab:ablation}).
    While closed-source frontier models like Gemini could theoretically generate these trees via in-context learning, deploying a 12B-parameter open-weights model addresses the critical robotic constraints of computational opacity and network latency ($P_4$). By distilling the structured task decomposition under constrained symbolic grammars of a massive foundation model into Pixtral-12B via constrained synthetic data, we yield an architecture capable of deterministic, low-latency, and entirely local execution on edge-deployable hardware, safeguarding proprietary factory environments from cloud dependency.

\subsection{Hardware and Implementation Details}
    Fine-tuning was executed on an \textbf{NVIDIA A40 GPU} with $48$ GiB of VRAM on remote cluster. Given the 12B parameter scale, we implemented a suite of memory-efficient strategies to enable single-GPU training:
    \begin{itemize}
        \item \textbf{Quantized Optimization:} We utilized the 8-bit Paged AdamW optimizer and Bfloat16 numerical precision to maintain stability while minimizing memory footprint.
        \item \textbf{Gradient Management:} We employed gradient checkpointing and a gradient accumulation strategy (steps $= 4$) to achieve an effective batch size of 4.
        \item \textbf{Context Window Constraints:} A significant challenge in training 12B-parameter multimodal models on single-GPU setups is the management of peak VRAM consumption during the backward pass. To ensure stable execution, we implemented a sequence length constraint. The primary model used in our physical experiments was trained with a maximum context window of 4,400 tokens. To investigate the impact of memory constraints on logical depth, we also trained variant models with reduced maximum token thresholds. Any training sample exceeding the 4,400 tokens limit was discarded to prevent \texttt{torch.OutOfMemoryError} failures.
    \end{itemize}

\subsection{Training Paradigm and Objective}
The model is trained via SFT to map an egocentric RGB image $I$, a system specification $S$, and a human instruction $U$ to a JSON-encoded BT $\mathcal{T}$. For each training sample $(I,S,U,\mathcal{T})$, the multimodal prompt and target tree are serialized into a single autoregressive sequence $X = (x_1,\dots,x_T)$.

Before training, we apply a deterministic preprocessing step to the target BT JSON. Specifically, all null-valued fields are removed recursively and the resulting object is serialized in minified form, i.e., without unnecessary whitespace or formatting tokens. This reduces the sequence length seen by the model while leaving the tree topology and all executable fields unchanged. 

Let \(\mathcal{P} \subseteq \{1,\dots,T\}\) denote the set of prompt-token positions and \(\mathcal{B} \subseteq \{1,\dots,T\}\) the set of target BT-token positions, with \(\mathcal{P} \cap \mathcal{B} = \emptyset\) and \(\mathcal{P} \cup \mathcal{B} = \{1,\dots,T\}\). To ensure strict adherence to the robotic symbolic schema and mitigate the risk of \textit{schema hallucination} during zero-shot execution, we deviate from standard instruction masking~\cite{huertaenochian2024instructionfinetuningdoesprompt}. Instead, we utilize a Weighted Prompt-Completion Loss $\mathcal{L}(\theta)$, defined over the entire sequence $X$:

\begin{equation}
\mathcal{L}(\theta) = - \sum_{t=1}^{T} \omega_t \log P(x_t | x_{<t}; \theta)
\end{equation}
where the per-token weight $\omega_t$ is defined as:
\begin{equation}
\omega_t =
\begin{cases}
\lambda_p, & \text{if } t \in \mathcal{P} \\
1, & \text{if } t \in \mathcal{B}
\end{cases}
\end{equation}

In our experiments, \(\lambda_p \in \{0,1\}\), corresponding respectively to standard completion-only masking and full-sequence supervision. The \(\lambda_p=1\) setting encourages the model to learn the boundary between the structured system specification and the symbolic BT output, which empirically improves schema compliance during zero-shot generation. This interpretation is consistent with the ablation results reported later in the paper, where prompt masking yields deceptively low training loss but inferior execution reliability.

\subsection{Training Evaluation Metrics}
    The primary objective during the SFT phase is the \textbf{minimization of the Causal Cross-Entropy Loss}, $\mathcal{L}$. However, to evaluate the model's mastery of the structured JSON-BT grammar, we also monitor the \textbf{Perplexity}. Perplexity provides a measure of how well the probability distribution predicted by the model aligns with the ground-truth robotic logic. Formally, for a target Behavior Tree sequence $Y = \{y_1, y_2, \dots, y_T\}$, the Perplexity is derived from the average negative log-likelihood:
    \begin{equation}
        PPL(Y) = \exp\left( \mathcal{L} \right)
    \end{equation}
    In our specific domain, a Perplexity value approaching $1.0$ indicates that the model has achieved a state of \textit{syntactic certainty}, where the structural tokens of the tree (e.g., \texttt{Sequence}, \texttt{Parallel}, and brackets) become highly predictable.
    
    We monitored training progress via four primary indicators: 
    \begin{itemize}
        \item \textbf{Causal Cross-Entropy Loss (LOSS)}: The standard objective for causal LLM tasks, measuring the discrepancy between the predicted token probability distribution and the ground-truth BT sequence during evaluation.
        \item \textbf{Perplexity (PPL)}: Provides an exponentiated measure of the loss, indicating how ”uncertain” the model is when generating the next part of a Behavior Tree. 
        \item \textbf{Peak of VRAM Usage (PVU)}: the maximum GPU memory consumed during the training of each run.
        \item \textbf{Training Latency (TL)}: How much time it takes to complete one pass over the dataset. This metric is provided to allow for a relative comparison of computational requirements across different model configurations. It should be noted that experiments were conducted on a shared performance computing cluster. Therefore, reported times reflect empirical observations in a non-isolated environment rather than an absolute measure of optimized system throughput.
    \end{itemize}
    Training behaviors of the models are reported in Tab.~\ref{tab:training} and discussed in Sec.~\ref{Sec: ablation}.

\subsection{Training Dynamics and Syntactic Convergence}
    To evaluate the efficacy of the fine-tuning process, we monitored the Causal Cross-Entropy Loss and Perplexity (PPL) during the supervised learning phase. As detailed in Tab. \ref{tab:training}, these metrics provide a quantitative measure of the model's syntactic certainty regarding the Behavior Tree JSON schema for the training process.

    We observe that a PPL threshold of approximately $1.85$ is required for consistent JSON validity (JVR). Models exceeding this threshold, such as A6 (PPL: $1.964$) and A15 (PPL: $1.865$), frequently suffered from structural problems (Sec.~\ref{Sec: ablation}), resulting in truncated trees or mismatched key-value pairs that failed the robotic parser.

\subsubsection{\textbf{Loss and Perplexity Analysis}}
    For the baseline configuration (A1), the PPL converged to $1.831$, representing a state where the structural tokens of the Behavior Tree (e.g., node types and bracket nesting) became highly predictable. In contrast, models with inadequate learning rates or restricted context windows exhibited significantly higher PPL. Specifically, A13 (High LR) reached a PPL of $18.916$, which directly correlated with a total collapse in generative coherence during inference. 
    
    The masked training objective ($\lambda_{p} = 0$) reached a near-zero convergence ($\mathcal{L} \approx 0.0016$) with minimal perplexity, suggesting a high degree of precision on the target completion. However, this metric proved deceptive; while the model attained high local precision, it failed to learn the global syntactic boundaries of the task (Sec.~\ref{Sec: Ablation}).

\subsubsection{\textbf{Hardware Constraints and Optimization}}
    The Peak VRAM Usage (PVU) and Training Latency (TL) metrics justify the selection of the Pixtral-12B backbone. As shown in Tab. \ref{tab:training}, the primary configuration A1 utilized $96.5\%$ of the available VRAM, defining $4400$ tokens as the hardware-constrained ceiling (discarding $130$ samples from the dataset). To maintain training stability and prevent \texttt{OutOfMemory} errors, we implemented a strict sample-exclusion policy: any training instance exceeding the context window was discarded from the dataset. Consequently, while A1 utilized the full $10,000$ sample distribution, restricted models like A6 were trained on a reduced, biased subset consisting primarily of low-complexity tasks.

\subsubsection{\textbf{Convergence and Dataset Scale}}
    The relationship between dataset scale and loss highlights the data-efficiency of the framework. We observed diminishing returns after $8,000$ samples (A17), where the PPL only improved by $0.015$ when scaling to $10,000$ samples. The constant improvement in increasing the number of samples suggests that while few samples are sufficient to learn the grammar of Behavior Trees, the final samples are critical for Logical Grounding, the precise mapping of natural language colors and geometries to the correct symbolic metadata IDs.

    \begin{table*}[t]
    \centering
    \begin{tabular}{c|ccccccc|cccc}
    \hline
    \textbf{ID} & \textbf{Module Map} & \textbf{Rank} & \textbf{Context (tokens)} & \textbf{Dropout}& \textbf{Learn. Rate} &\textbf{Dataset Size} & $\mathbf{\lambda_p}$ &\textbf{LOSS} & \textbf{PPL} & \textbf{PVU (\%)}& \textbf{TL (h)} \\ \hline
    \rowcolor{blue!20}A1 & \texttt{heavy} & 16 & 4400 & 0.05 & $1\cdot10^{-4}$ & 10k & 1 & 0.605 & 1.831& 96.506 & 15.12\\
    \hline
    A2 & \cellcolor{blue!10}\texttt{medium} & 16 & 4400 & 0.05 & $1\cdot10^{-4}$ & 10k & 1 & 0.615 & 1.850 & 98.580 & 12.92\\
    \rowcolor{gray!10}A3 & \cellcolor{blue!20}\texttt{lite} & 16 & 4400 & 0.05 & $1\cdot10^{-4}$ & 10k & 1 &  0.6203 & 1.859 & 98.411 & 12.59\\
    \hline
    A4 & \texttt{heavy} & \cellcolor{blue!10}8 & 4400 & 0.05 & $1\cdot10^{-4}$ & 10k & 1 & 0.608 & 1.837 & 98.529 & 15.18\\
    \rowcolor{gray!10}A5 & \texttt{heavy} & \cellcolor{blue!20}32 & 4400 & 0.05 & $1\cdot10^{-4}$ & 10k & 1 & 0.6117 & 1.844 & 95.741 & 15.12\\
    %A6 & \texttt{heavy} & \cellcolor{blue!10}64 & 4400 & 0.05 & $1\cdot10^{-4}$ & 10k & / & / & out of memory & / \\
    \hline
    A6 & \texttt{heavy} & 16 & \cellcolor{blue!10}1200 & 0.05 & $1\cdot10^{-4}$ & 10k & 1 & 0.675 & 1.964 & 66.979 & 8.41\\
    \rowcolor{gray!10}A7 & \texttt{heavy} & 16 & \cellcolor{blue!20}2400 & 0.05 & $1\cdot10^{-4}$ & 10k & 1 & 0.616 & 1.852 & 78.528 & 13.98\\
    A8 & \texttt{heavy} & 16 & \cellcolor{blue!10}3600 & 0.05 & $1\cdot10^{-4}$ & 10k & 1 & 0.606 & 1.833 & 92.073 & 15.02\\
    \rowcolor{gray!10}A9 & \texttt{heavy} & 16 & \cellcolor{blue!20}4200 & 0.05 & $1\cdot10^{-4}$ & 10k & 1 & 0.605 & 1.831 & 96.506 & 15.12\\
    %A11 & \texttt{heavy} & 16 & \cellcolor{blue!10}4600 & 0.05 & $1\cdot10^{-4}$ & 10k & / & / & out of memory & / \\
    \hline
    A10 & \texttt{heavy} & 16 & 4400 & \cellcolor{blue!10}0.0 & $1\cdot10^{-4}$ & 10k & 1 & 0.607 & 1.835 & 96.158 & 14.51\\
    \rowcolor{gray!10}A11 & \texttt{heavy} & 16 & 4400 & \cellcolor{blue!20}0.1 & $1\cdot10^{-4}$ & 10k & 1 & 0.608 & 1.837 & 96.506 & 15.11\\
    A12 & \texttt{heavy} & 16 & 4400 & \cellcolor{blue!10}0.2 & $1\cdot10^{-4}$ & 10k & 1 & 0.608 & 1.837 & 96.506 & 15.12\\
    \hline
    \rowcolor{gray!10}A13 & \texttt{heavy} & 16 & 4400 & 0.05 & \cellcolor{blue!20}$1\cdot10^{-3}$ & 10k & 1 & 2.940 & 18.916 & 96.506 & 15.09\\
    A14 & \texttt{heavy} & 16 & 4400 & 0.05 & \cellcolor{blue!10}$1\cdot10^{-5}$ & 10k & 1 & 0.637 & 1.891 & 96.506 & 15.05\\
    \hline
    \rowcolor{gray!10}A15 & \texttt{heavy} & 16 & 4400 & 0.05 & $1\cdot10^{-4}$ & \cellcolor{blue!20}2k & 1 & 0.623 & 1.865 & 95.980 & 2.70\\
    A16 & \texttt{heavy} & 16 & 4400 & 0.05 & $1\cdot10^{-4}$ & \cellcolor{blue!10}5k & 1 & 0.617 & 1.853 & 96.457 & 7.13\\
    \rowcolor{gray!10}A17 & \texttt{heavy} & 16 & 4400 & 0.05 & $1\cdot10^{-4}$ & \cellcolor{blue!20}8k & 1 & 0.613 & 1.846 & 96.457 & 11.82\\ \hline
    A18 & \texttt{heavy} & 16 & 4400 & 0.05 & $1\cdot10^{-4}$ & 10k & \cellcolor{blue!10}0 & 0.0016 & 1.001& 96.458 & 15.45 
    \\
    \hline
    \end{tabular}

    \vspace{4pt}
    \caption{Training Metrics and Resource Utilization on NVIDIA A40. The reported model A1 is the one used for real-world tests and SOTA comparison.}
    \label{tab:training}
    \end{table*}

\section{Behavior Trees Generation with Trained Models}\label{Sec: Ablation}
    This section details the inference pipeline and evaluates the performance of the fine-tuned Pixtral-12B variants in synthesizing Behavior Trees (BTs). We focus on the models' ability to map high-dimensional visual features to precise symbolic structures while adhering to the atomic primitive constraints defined in Sec.~\ref{Sec: problem and solution}.

\subsection{Cross-Domain Sim-to-Real Transfer and Visual Robustness}
    A primary challenge in robot learning is the domain gap between synthetic training environments and real-world settings. To evaluate the robustness of our approach, all qualitative and quantitative \textbf{experiments} discussed in this work \textbf{were conducted using real-world egocentric images}, despite the model being trained exclusively on synthetic MuJoCo data. This confirms that the proposed pipeline achieves a Zero-Shot Logic Transfer, where the model generalizes from idealized synthetic spatial relations to physical environments without intermediate real-world fine-tuning in target domain.

\begin{figure}[b]
       \centering
       \includegraphics[width=0.49\linewidth]{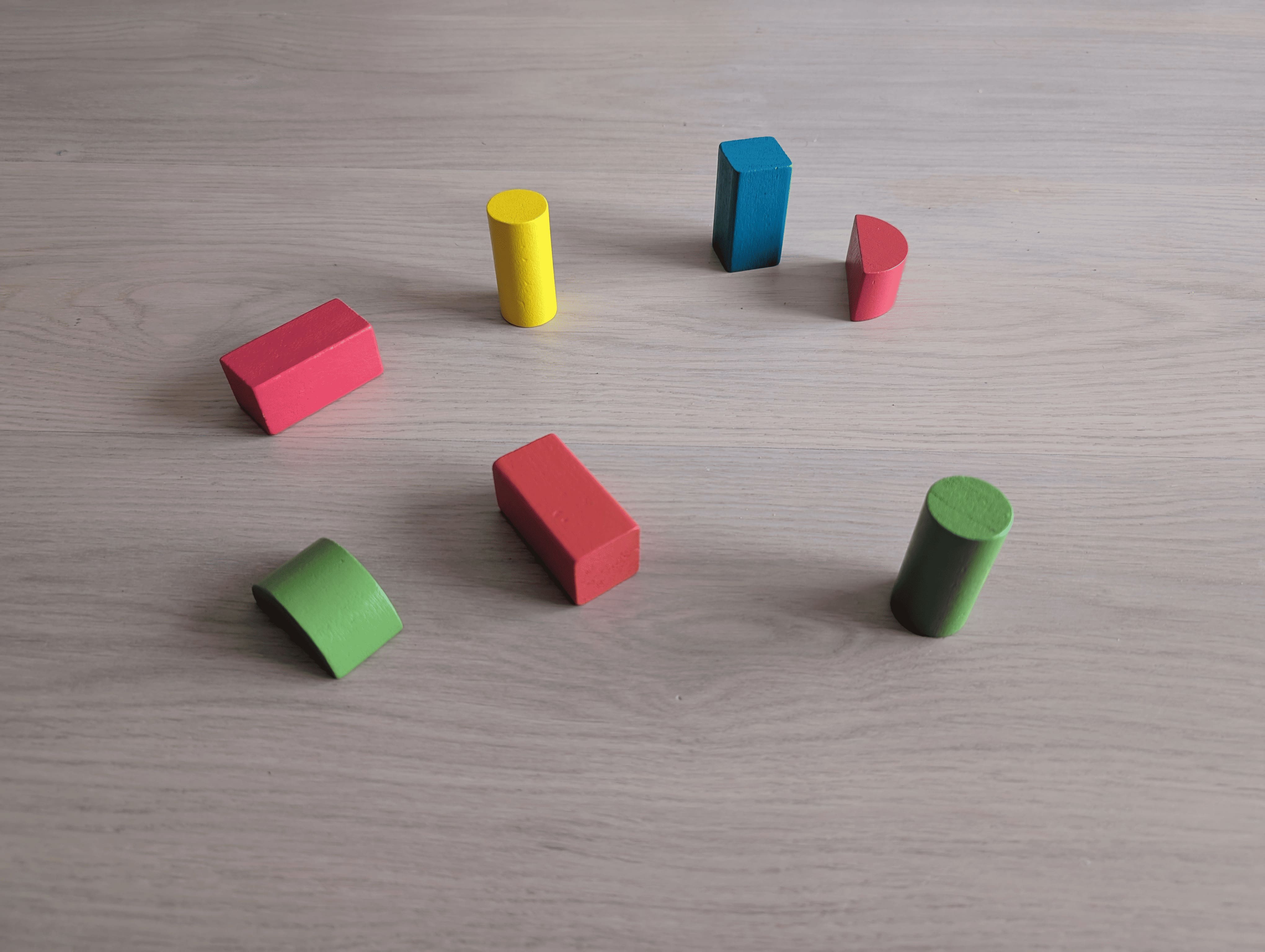}
       \includegraphics[width=0.49\linewidth]{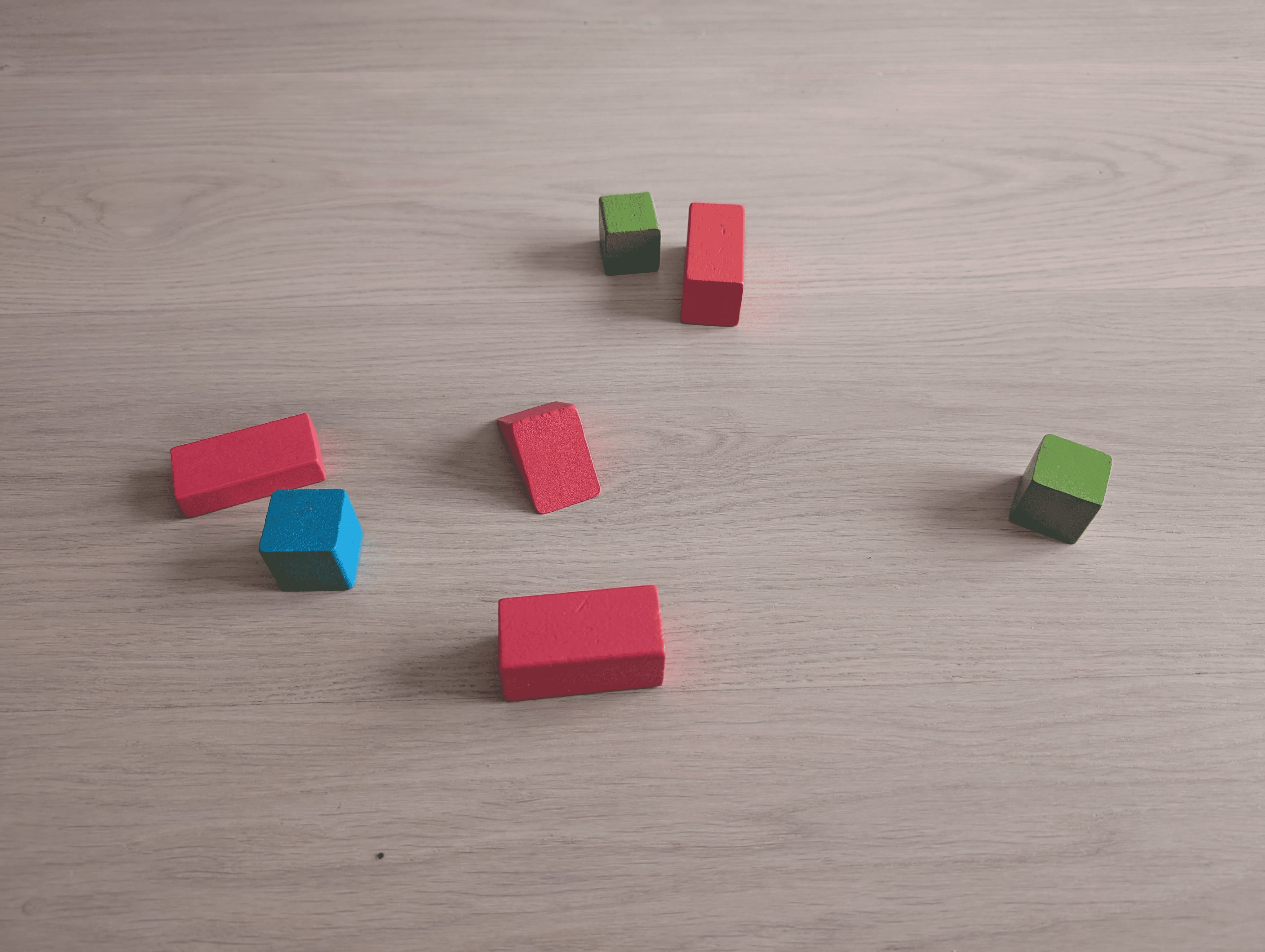}
       \caption{Example of real-world images, representing scenarios coherent with the synthetic dataset.}
       \label{fig: ablation images}
   \end{figure}

    As demonstrated by the results in Section~\ref{Sec: ablation}, the fine-tuned Pixtral-12B exhibited remarkable \textbf{invariance between real} (Fig.~\ref{fig: ablation images}) \textbf{and synthetic images} (Fig.~\ref{fig: synthetic images}). 
    By exposing the model to high variance in object placement within simulation, the model effectively learned a robust mapping between high-level symbolic labels and their corresponding visual features.

    Crucially, the model maintained its ability to perform \textbf{Referential Grounding}, such as distinguishing between "left" and "right" objects (Sec.~\ref{sec: visual referential}), when presented with real-world textures, shadows, and varying lens distortions that were absent during training. This suggests that the LoRA adaptation of the MLP layers does not merely memorize the low-level visual patterns of the MuJoCo renderer. Instead, it internalizes the spatial logic required for task planning. This effective bridging of the Sim-to-Real gap proves that \textbf{procedurally generated Behavior Tree datasets can serve as high-fidelity proxies for real-world robotic reasoning}, significantly reducing the need for costly manual annotation of real-world datasets.

\subsection{Inference Pipeline and Prompting}
    To evaluate the efficacy of the fine-tuning process, we developed an automated inference pipeline that compares the base Pixtral-12B model against various LoRA adapters. For each test case, the model is provided with an RGB image $I$, the user request $U$, and a system specification $S$ containing dynamic metadata and the primitive library. 

    We utilize greedy decoding ($T=0$) during generation to ensure deterministic and reproducible results, which is critical for safety-verifiable robotics. The model is prompted to generate the assistant response following the chat template defined during training. To manage computational overhead, we set a generation limit of 4,400 new tokens. In our comparative tests, the base model always failed to adhere to the JSON schema and frequently produced linear sequences without reactivity, whereas the fine-tuned adapters consistently generated structured trees that are executed and task effective.

\subsection{Quantitative Performance Metrics}
    To rigorously assess the fine-tuned models, we employ a multi-layered evaluation framework spanning syntactic integrity and behavioral logic.
    \begin{itemize}
        \item \textbf{Structural Complexity and Syntactic Integrity:}
        \begin{enumerate}
            \item \textbf{Task Success Rate (TSR)}: Physical success of the Behavior Tree. Success is manually evaluated on task execution only for BTs with JSON validity and no Key Errors. In this context, task success implicitly reflects both perceptual grounding accuracy and structural validity of the generated Behavior Trees.
            \item \textbf{JSON Validity Rate (JVR)}: The percentage of model outputs that are successfully parsed by a standard JSON library.
            \item \textbf{Key Error Rate (KER)}: Frequency of missing mandatory fields (\texttt{type, args, children, name}) required by the BT parser or additional not required fields (\texttt{output, usage}). If the JSON is not valid, the metric can not be computed.
            \item \textbf{Tree Depth (TD)}: Maximum nesting levels, representing logical complexity.
            \item \textbf{Leaf Count (LC)}: Total number of terminal Actions and Conditions.
            \item \textbf{Node Density (ND)}: Ratio of composite nodes to leaf nodes.
            \item \textbf{Mean Execution Nodes (MEN)}: The average number of nodes generated per successful task.
        \end{enumerate}
        \item \textbf{Behavioral Logic and Reactivity:}
        \begin{enumerate}
            \item \textbf{Schema Compliance (SC)}: Adherence to task-specific rules (e.g., approach phases).
            \item \textbf{Inference Latency (IL)}: Average time to generate a complete tree. Reported values represent the mean execution time across the evaluation set. Given that measurements were obtained in a shared performance computing environment, these figures serve as a representative benchmark of the model's computational complexity rather than an absolute measure of real-time performance on dedicated robotic edge hardware. Our VLM operates asynchronously as an offline policy compiler. While the generation of the Behavior Tree takes $\simeq60$ seconds, the resulting JSON is a lightweight, reactive structure. Once deployed to the robot's local execution engine, the BT runs at high frequency, enabling instantaneous reactive guarding and recovery without querying the VLM again.
        \end{enumerate}
    \end{itemize}

\subsection{Ablation Study}\label{Sec: ablation}
   To identify the optimal configuration for multimodal Behavior Tree synthesis, we conducted an extensive ablation study across 18 model variants. Notably, this analysis strictly evaluates the synthesis of structured symbolic policies and excludes comparisons with end-to-end VLA models. Because VLA models output high-dimensional action chunks rather than discrete logical steps, a direct quantitative comparison with our interpretable BT generation framework is structurally incongruent.
   We evaluate these models based on training convergence (Tab.~\ref{tab:training}) and inference performance (Tab.~\ref{tab:ablation}) across 20 new real-world test cases. 
   
\subsubsection{\textbf{Impact of Module Mapping and Capacity}}
    The transition from a \textbf{light} configuration (A3), targeting only attention projections, to a \textbf{heavy} configuration (A1), which incorporates MLP blocks (\texttt{gate\_proj, up\_proj, down\_proj}), reveals a fundamental requirement for symbolic reasoning. As shown in Tab.~\ref{tab:ablation}, A3 fails almost entirely (10\% TSR), often defaulting to natural language descriptions instead of JSON. Conversely, the inclusion of MLP layers in A1 and A2 enables the model to internalize the rigid JSON-BT grammar. While A2 (medium) achieves 100\% JVR, it suffers from a 60\% Key Error Rate (KER), frequently mislabeling node types. This confirms that \textit{heavy} module adaptation is necessary to map visual-spatial relationships to the correct symbolic categories.

\subsubsection{\textbf{LoRA Rank and Expressivity}}
    We investigated the impact of the LoRA rank $r \in \{8, 16, 32\}$. A lower rank ($r=8$, A4) results in "syntactic fragility," with a JVR of only 50\% due to unclosed brackets and truncated hierarchies. Interestingly, increasing the rank to $r=32$ (A5) does not yield linear improvements; while JVR remains high, the KER rises to 30\% due to structural "noise" and hallucinated node arguments. The results suggest that $r=16$ provides the optimal balance of expressive power for hierarchical nesting without overfitting to specific training templates.

\subsubsection{\textbf{Context Window and VRAM Constraints}}
    The context window size dictates the maximum complexity of the robotic logic the model can internalize. As shown in Tab. \ref{tab:ablation}, a window of 1,200 tokens (A6) leads to a total failure in multi-object tasks (0\% TSR). This is not merely a result of inference-time truncation, but of \textit{dataset filtering} during training; at 1,200 tokens, nearly all long-horizon manipulation samples were excluded from the training set. 

    As a result, the model trained under A6 was only exposed to simple "Lift" and "Move" primitives. During inference, when presented with a complex "Stack" command, the model experienced a catastrophic distribution shift, unable to synthesize the recursive structures it was never permitted to observe during training. Only by increasing the window to 4,400 tokens (A1) (thereby preserving the full diversity of the 10,000 sample dataset) was the model able to learn the high-level coordination required for multi-stage reactive planning.
\subsubsection{\textbf{Learning Rate and Convergence Stability}}
    The learning rate proved to be the most sensitive hyperparameter. A rate of $10^{-3}$ (A13) causes catastrophic forgetting, resulting in nonsensical character loops (PPL 18.916). Conversely, $10^{-5}$ (A14) is insufficient to adapt the model to the JSON schema, leading to a 20\% TSR. 
    
    Regarding regularization, a dropout of 0.05 (A1) provides the best generalization. Increasing dropout to 0.2 (A12) interferes with the deterministic precision required for JSON syntax, dropping TSR to 30\% due to "shuffled" keys and inconsistent node labeling.

\subsubsection{\textbf{Dataset Scaling}}
    Finally, we analyzed the data efficiency of the pipeline (A15--A17). At 2,000 samples, the model learns the concept of JSON but lacks spatial grounding. The most significant jump occurs between 2,000 and 5,000 samples (TSR 10\% to 50\%), where the model begins to handle multi-object logic. Scaling to 10,000 samples (A1) provides the final layer of robustness needed for 100\% success, particularly in resolving referential ambiguity (e.g., "leftmost object") across diverse scenes.
    
    Overall, these results suggest that the model does not simply memorize output templates but learns structural regularities of robotic policies, as evidenced by stable syntactic validity across configurations and robustness to variations in scene composition.

    \begin{table*}[t]
    \centering
    \begin{tabular}{c|ccc|cccc|ccc}
    \hline
    \textbf{ID} &\textbf{TSR (\%)} &\textbf{JVR (\%)} & \textbf{KER (\%)} & \textbf{TD}& \textbf{LC}& \textbf{ND}& \textbf{MEN} & \textbf{SC (\%)} &\textbf{IL (s)} \\ \hline
    GPT-5 zero-shot & 0 & 0 & 0 & 0 & 0 & 0 & 0 & 0 & N.A.\\
    \rowcolor{gray!10}GPT-5 one-shot & 80 & 80 & 0 & 6.2 & 40.4 & 0.58 & 64.0 & 80 & N.A.\\
    Gemini 3 Flash zero-shot & 0 & 0 & 0 & 0 & 0 & 0 & 0 & 0 & N.A.\\
    \rowcolor{gray!10}Gemini 3 Flash one-shot & 100 & 100 & 0 & 6.2 & 33.6 & 0.56 & 52.2 & 100 & N.A.\\
    \hline
    Pixtral-12B & 0 & 0 & 0 & 0 & 0 & 0 & 0 & 0 & 52.80\\
    \hline
    \rowcolor{blue!20}A1 & 100 & 100 & 0 & 6.0 & 23.1 & 0.59 & 30.1 & 100 & 61.85 \\ \hline
    A2 & 40 & 100 & 60 & 6.0 & 12.8 & 0.30 & 33.4 & 100 & 58.45 \\
    \rowcolor{gray!10}A3 & 10 & 10 & 0 & 0.6 & 1.9 & 0.06 & 3.0 & 10 & 53.86 \\
    \hline
    A4 & 20 & 50 & 0 & 3.4 & 11.2 & 0.28 & 15.6 & 100 & 54.66 \\
    \rowcolor{gray!10}A5 & 70 & 100 & 30 & 6.0 & 21.8 & 0.48 & 32.1 & 90 & 54.42 \\
    \hline
    A6 & 0 & 10 & 0 & 0.8 & 2.1 & 0.04 & 3.5 & 0 & 18.42 \\
    \rowcolor{gray!10}A7 & 30 & 40 & 0 &2.1 & 6.7 & 0.2 & 10.1 & 30 &40.54 \\
    A8 & 40 & 100 & 60 & 6.0 & 12.8 & 0.30 & 33.4 & 100 & 61.64 \\
    \rowcolor{gray!10}A9 & 90 & 100 & 0 & 6.0 & 22.1 & 0.54 & 35.8 & 100 & 60.12 \\
    \hline
    A10 & 80 & 100 & 10 & 6.0 & 25.1 & 0.58 & 38.4 & 90 & 52.24 \\
    \rowcolor{gray!10}A11 & 60 & 90 & 20 & 5.4 & 19.2 & 0.42 & 28.1 & 100 & 54.39 \\
    A12 & 30 & 70 & 30 & 4.2 & 14.5 & 0.31 & 19.8 & 80 & 54.45 \\
    \hline
    \rowcolor{gray!10}A13 & 0 & 0 & 0 & 0 & 0 & 0 & 0 & 0 & 309.71 \\
    A14 & 20 & 40 & 10 & 2.4 & 7.1 & 0.18 & 12.4 & 40 & 58.27 \\
    \hline
    \rowcolor{gray!10}A15 & 10 & 40 & 20 & 1.8 & 5.2 & 0.12 & 8.4 & 20 & 55.45 \\
    A16 & 50 & 90 & 10 & 4.8 & 15.6 & 0.38 & 24.5 & 80 & 58.21 \\
    \rowcolor{gray!10}A17 & 70 & 100 & 0 & 6.0 & 21.4 & 0.51 & 34.1 & 100 & 59.88 \\
    \hline
    A18 & 0 & 100 & 100 & 0 & 0 & 0 & 0 & 100 & 138.55 \\
    \hline
    \end{tabular}

    \vspace{4pt}
    \caption{Ablation Study: Impact of Hyperparameters on BT Synthesis Quality. All the models are on 20 different user requests $U$ that are coherent with the training set (in distribution scenarios) and with real-world $I$, \textbf{despite zero real-world training samples}. A1-18 are fine-tuned specialists (10k synthetic training examples). GPT-5 and Gemini 3 Flash are prompted generalists with no task-specific training. Results are not directly comparable but illustrate the trade-off between fine-tuning cost and zero-shot structural compliance.}
    \label{tab:ablation}
    \end{table*}

\subsection{Comparative Analysis with SOTA Models: fine-tuned specialist vs. prompted generalist}\label{sec: SOTA}
    To contextualize the performance of the fine-tuned specialist model, we compare it against two state-of-the-art frontier models, \textbf{GPT-5} and \textbf{Gemini 3 Flash}, under zero-shot and one-shot prompting regimes. We emphasize that this comparison is structurally asymmetric by design: our model has been fine-tuned on 10,000 in-distribution synthetic examples and therefore possesses task-specific structural priors, while the frontier models receive no task-specific training. The comparison is not intended to demonstrate superiority over larger models, but rather to evaluate two complementary questions: whether generalist frontier models can reliably perform constrained symbolic BT synthesis without fine-tuning, and whether a compact, locally-deployable specialist can match their output quality within the targeted manipulation domain. This distinction is practically relevant, as frontier models require cloud access, introduce latency and privacy constraints, and consume significantly more inference-time compute, all of which are prohibitive for real-time robotic edge deployment.

    \subsubsection{\textbf{Zero-Shot Performance and Structural Divergence}}
    In the initial Zero-Shot regime, the models were provided with the image $I$, the instruction $U$, and the baseline system specification $S$ (primitive library and metadata) as the fine-tuned models. Under these conditions, both SOTA models failed to produce executable outputs. Instead of the required JSON-BT schema, the models defaulted to natural language descriptions or "text-based" tree visualizations. 
    
    When explicitly prompted to enforce a JSON output format, the models exhibited significant \textit{schema misalignment}. Specifically, the SOTA models frequently hallucinated non-functional keys, such as assigning a \texttt{"name"} field to composite nodes (e.g., \texttt{Sequence, Selector}), a structure that is not supported by our symbolic execution engine and results in immediate parser-level failure. This behavior indicates that while foundation models possess high-level task-planning capabilities, they lack the inherent structural discipline required for zero-shot deployment in strictly constrained robotic environments.

    \subsubsection{\textbf{One-Shot In-Context Learning}}
    To facilitate a meaningful comparison of logical reasoning, we modified the system specification $S$ to include a single \textbf{positive example} of a valid JSON-encoded Behavior Tree (One-Shot prompting) as did in Sec.~\ref{Sec: multimodal prompt geenration} for meaningful dataset generation. Under this regime, both GPT-5 and Gemini 3 Flash succeeded in adhering to the syntactic constraints of our framework with a high percentage. However, in some cases with the GPT-5 model, the JSON format is not respected at all, with the addition of unpredictable artifacts that compromise the execution of the associated BT while running the dedicated script. Instead, Gemini achieves a 100\% score in both \textbf{TSR} and \textbf{JVR} for producing structured text, justifying the model choice for dataset construction.
    As shown in Tab. \ref{tab:ablation}, within the targeted manipulation domain and under zero-shot conditions, \textbf{A1} matches the output quality of one-shot-prompted frontier models, confirming that fine-tuning cost can substitute for in-context examples in structurally constrained settings.
    This highlights the primary advantage of our approach: through synthetic data supervision, our model has internalized the robotic grammar as an intrinsic prior. Unlike SOTA models that require precious context-window space to be occupied by one-shot examples to maintain structural validity, our trained model performs zero-shot synthesis with $100\%$ syntactic reliability within the targeted manipulation domain, preserving the full context for task-relevant spatial reasoning and metadata.

    \subsection{The Impact of Syntactic Anchoring on Reliability}
    Our empirical results revealed a counterintuitive relationship between training loss and execution reliability. Models trained with standard prompt masking ($\lambda_{p}=0$) achieved a near-zero training loss ($\mathcal{L} \approx 0.0016$), yet exhibited significant \textit{Functional Grounding Gaps} during inference (A18). These models frequently failed to terminate sequences or "backslid" into the verbose formatting style of the input specifications (e.g., hallucinating \textit{"usage"} and \textit{"output"} keys into the symbolic tree).

    Conversely, the proposed Anchored Loss ($\lambda_{p}=1$, yielding $\mathcal{L} \approx 0.6$) demonstrated 100\% schema compliance. These observations align with recent theoretical work in~\cite{huertaenochian2024instructionfinetuningdoesprompt}, which suggests that including the prompt in the loss objective serves as a crucial regularizer. We hypothesize that this \textit{Syntactic Anchoring} forces the model to learn the \textit{Negative Constraints} of the task. Specifically, the model learns which metadata attributes are strictly input-bound and must be excluded from the generated execution trace. This boundary learning provides the necessary regularization to ensure robust, zero-shot performance in real-world robotic environments where strict schema adherence is safety-critical.

\section{Real-World Robotic Validation}\label{Sec: real-world}
    To evaluate the practical applicability and generalization of the VLM-synthesized Behavior Trees, we deployed the fine-tuned \textbf{Model A1} across two distinct industrial robotic platforms: a 7-DOF \textbf{Franka Emika Panda} and a 6-DOF \textbf{Universal Robots UR5e}, as all tested in-distribution evaluation scenarios were completed successfully (Tab.~\ref{tab:ablation}). The primary objective is to demonstrate that the high-level task plans synthesized by the VLM are platform-agnostic and robust to the visual domain gap between MuJoCo simulation and real-world environments. By maintaining a strict abstraction layer between the symbolic BT synthesis and the low-level hardware drivers, we verify that the model has internalized universal robotic logic, such as guarding and spatial offsetting, rather than platform-specific trajectories.

\subsection{Visual Robustness Under Sim-to-Real Transfer}
    In order to evaluate the visual invariance of the model, we established an \textbf{identical physical setup for both robotic hardware} to ensure a fair comparison. The workspace featured a tabletop environment with geometric primitives (cubes, cylinders, and parallelepipeds) randomized in color and position, mirroring the MuJoCo training distribution. For each inference cycle, we utilized \textbf{raw real-world egocentric images} captured from a top-down camera. This required the model to perform zero-shot visual grounding, identifying physical objects and their spatial relationships directly from the RGB feed without any synthetic pre-processing.

\subsection{Hardware-Agnostic Planning via Action Abstraction}
\begin{figure}[h]
        \centering
        \includegraphics[width=0.48\linewidth]{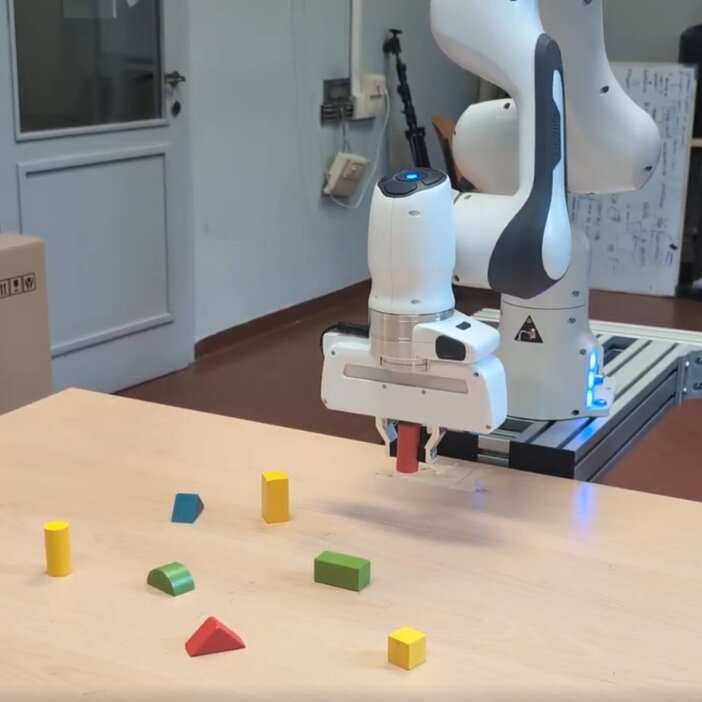}
        \includegraphics[width=0.48\linewidth]{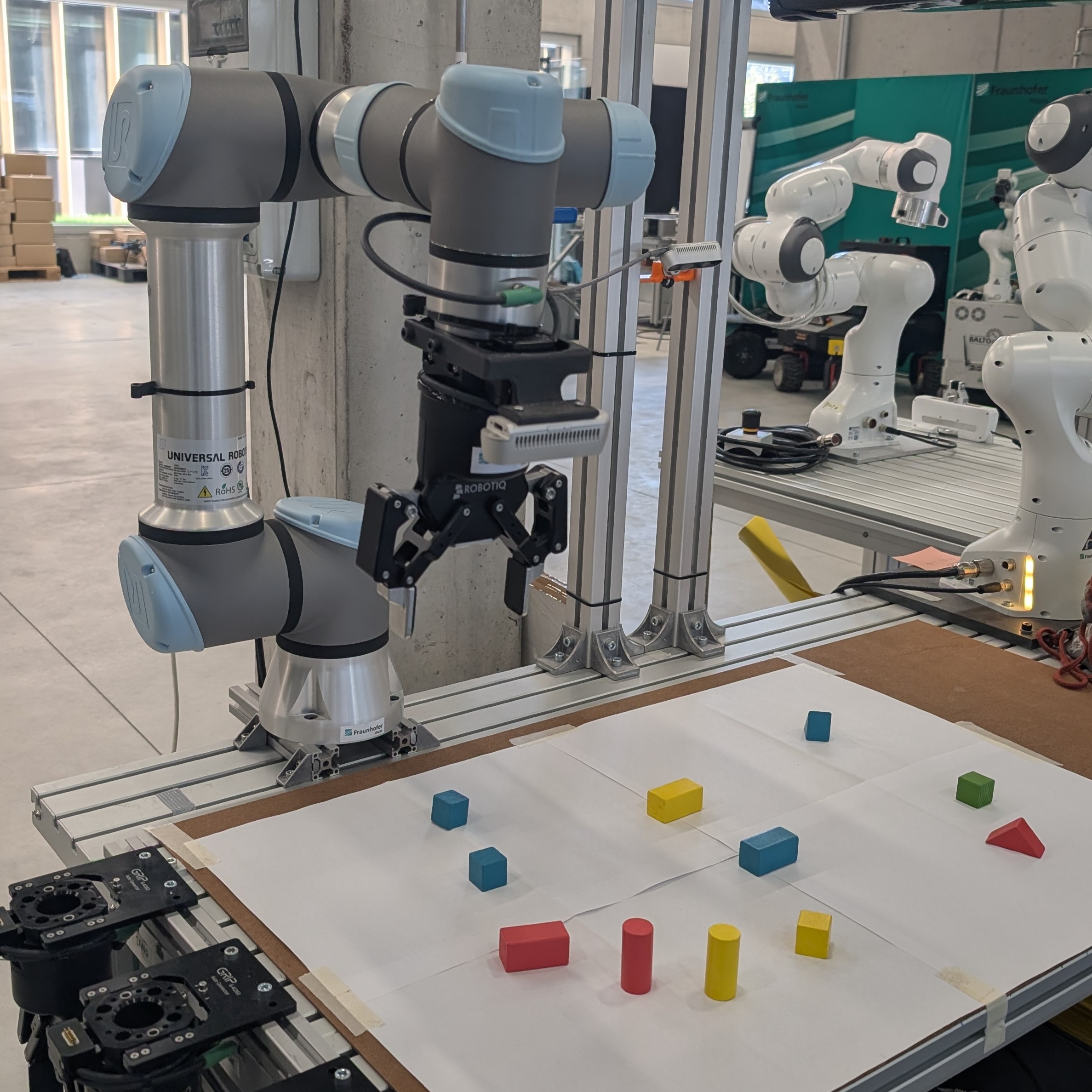}
        \caption{Real-world experimental platforms used to validate hardware-agnostic BT execution: the Franka Emika Panda and the UR5e.}
        \label{fig:robots}
    \end{figure}

    The proposed framework is \textbf{hardware-agnostic} at the planning level. By representing manipulation skills through the atomic primitive library of Sec.~\ref{Sec: Atomic Actions}, the VLM synthesizes BTs over symbolic actions and reactive constraints rather than robot-specific control details. As a result, the generated plan is \textbf{independent} of inverse kinematics, joint-space control, and end-effector implementation. Any platform exposing the same primitive interface can therefore execute the synthesized BT without modification, enabling a single fine-tuned model to generalize across heterogeneous manipulators (Fig.~\ref{fig:robots}).
        
\subsection{Results and Physical Execution}
    We conducted multiple trials for each robot across a suite of pick-and-place and stacking instructions. For all requests aligned with the semantic scope of the training data, the synthesized BTs (e.g., the BT reported in Fig.~\ref{fig: example BT print} of which execution is reported in Fig.~\ref{fig: execute printed BT}) were \textbf{successfully executed on both robots} without failure in all the tested cases.
    
    Quantitatively, the robots achieved consistent success across repeated runs. We observed that the reactive nature of the generated trees was vital for reliability; for instance, the robots would autonomously re-evaluate condition nodes (e.g., \texttt{is\_at\_pose}) before proceeding to the next atomic action, effectively managing the slight mechanical variances between the Panda and the UR5e. These results demonstrate that high-fidelity synthetic logic supervision is sufficient to generate robust, portable plans that \textbf{transfer seamlessly to physical hardware}.

    The dual-platform success confirms a critical finding: Model A1, \textbf{despite having never observed} a physical robot or\textbf{ a real-world photo} during its training phase, \textbf{internalizes a logic representation} robust enough to ignore the visual domain gap and mechanical variances of different hardware.
    
\begin{figure}[h]
    \centering
    
    \begin{tikzpicture}[placeholder/.style={draw=gray!30, fill=white, rounded corners=1pt, font=\fontsize{6}{7}\selectfont, align=center},
    arrow/.style={-{Stealth[scale=0.8]}, thick, draw=gray!60},
    % Group Styles
    group_bg/.style={draw=gray!15, fill=#1, rounded corners=3pt, inner sep=3pt}]

    \node[placeholder, xshift=0.1cm, 
      font=\fontsize{5}{6}\selectfont\ttfamily, 
      inner sep=2pt, align=left, draw=green!40] (thumb_bt) {
    \textbf{Behavior Tree:}\\
    \btS{[o]} Selector\\
    \hspace{10pt}\btA{$\rightarrow$} ? is\_at\_home\\
    \btS{\{-\}} Sequence\\
    \hspace{10pt}\btS{[o]} Selector\\
    \hspace{20pt}\btA{$\rightarrow$} ? is\_grasped\\
    \hspace{20pt}\btS{\{-\}} Sequence\\
    \hspace{30pt}\btS{[o]} Selector\\
    \hspace{40pt}\btA{$\rightarrow$} ? is\_gripper\_open\\
    \hspace{40pt}\btA{$\rightarrow$} OpenGripper\\
    \hspace{30pt}\btS{/\_/} Parallel\\
    \hspace{40pt}\btA{$\rightarrow$} ? is\_at\_pose(yellow\_parallelepiped\_1 + z\_offset)\\
    \hspace{40pt}\btA{$\rightarrow$} MovePose(yellow\_parallelepiped\_1 + z\_offset)\\
    \hspace{30pt}\btS{/\_/} Parallel\\
    \hspace{40pt}\btA{$\rightarrow$} ? is\_at\_pose(yellow\_parallelepiped\_1)\\
    \hspace{40pt}\btA{$\rightarrow$} MovePose(yellow\_parallelepiped\_1)\\
    \hspace{30pt}\btA{$\rightarrow$} CloseGripper\\
    \hspace{30pt}\btS{/\_/} Parallel\\
    \hspace{40pt}\btA{$\rightarrow$} ? is\_at\_pose(yellow\_parallelepiped\_1 + z\_offset)\\
    \hspace{40pt}\btA{$\rightarrow$} MovePose(yellow\_parallelepiped\_1 + z\_offset)\\
    \hspace{10pt}\btS{/\_/} Parallel\\
    \hspace{20pt}\btA{$\rightarrow$} ? is\_at\_pose(blue\_parallelepiped\_2 + z\_offset)\\
    \hspace{20pt}\btA{$\rightarrow$} MovePose(blue\_parallelepiped\_2 + z\_offset)\\
    \hspace{10pt}\btS{/\_/} Parallel\\
    \hspace{20pt}\btA{$\rightarrow$} ? is\_contact\\
    \hspace{20pt}\btA{$\rightarrow$} MoveDown\\
    \hspace{10pt}\btA{$\rightarrow$} OpenGripper\\
    \hspace{10pt}\btS{/\_/} Parallel\\
    \hspace{20pt}\btA{$\rightarrow$} ? is\_at\_pose(blue\_parallelepiped\_2 + z\_offset)\\
    \hspace{20pt}\btA{$\rightarrow$} MovePose(blue\_parallelepiped\_2 + z\_offset)\\
    \hspace{10pt}\btA{$\rightarrow$} home
};
        
    \end{tikzpicture}
    
    \caption{Example of BT obtained asking to the trained model to "Stack the yellow parallepiped on top of the blue parallelepiped".}
    \label{fig: example BT print}
\end{figure}

\begin{figure}
    \centering
    \includegraphics[width=0.24\linewidth]{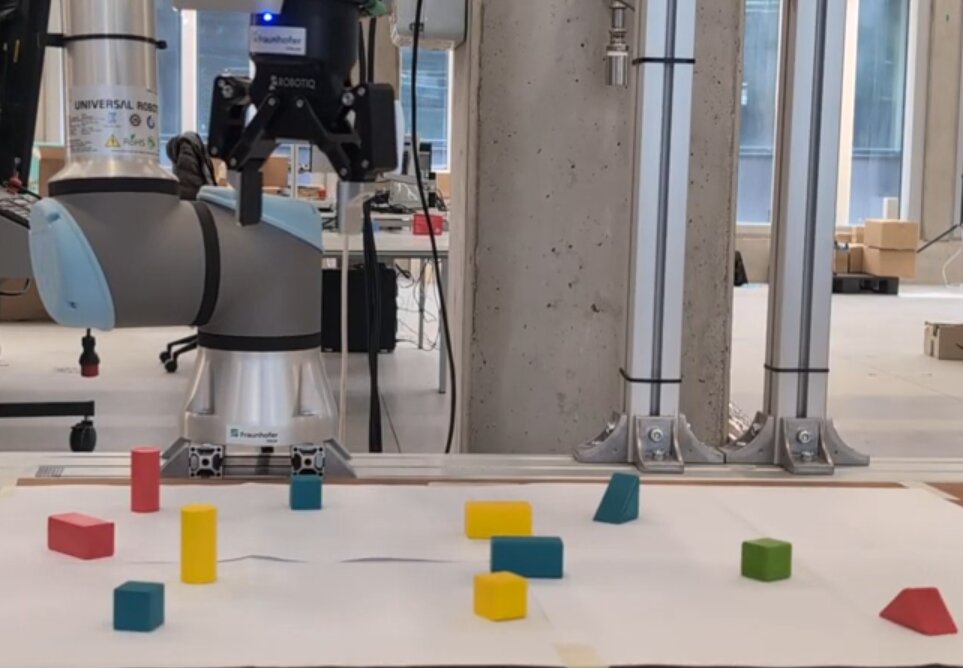}
    \includegraphics[width=0.24\linewidth]{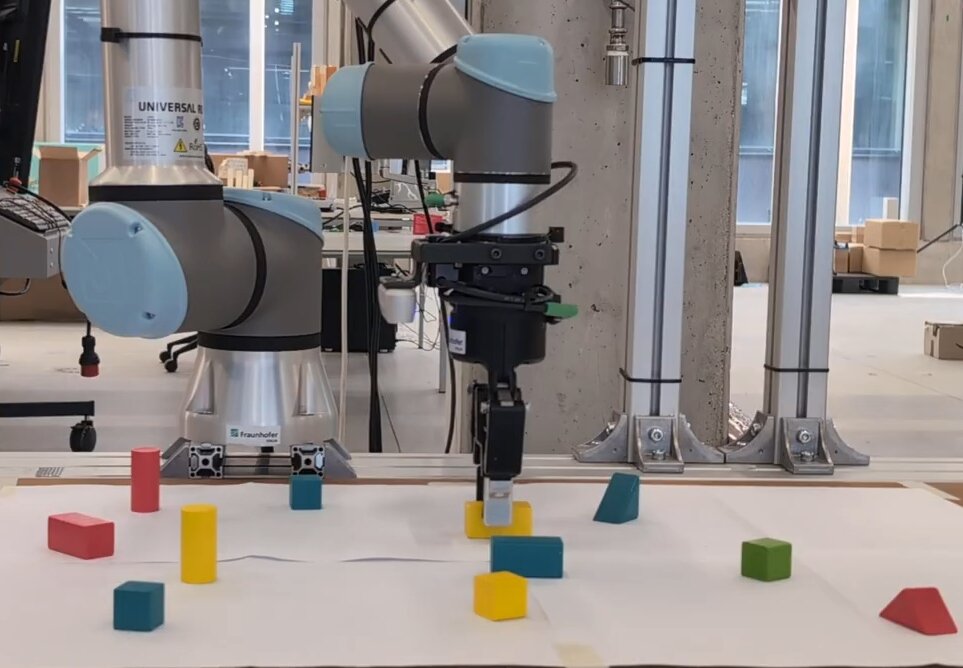}
    \includegraphics[width=0.24\linewidth]{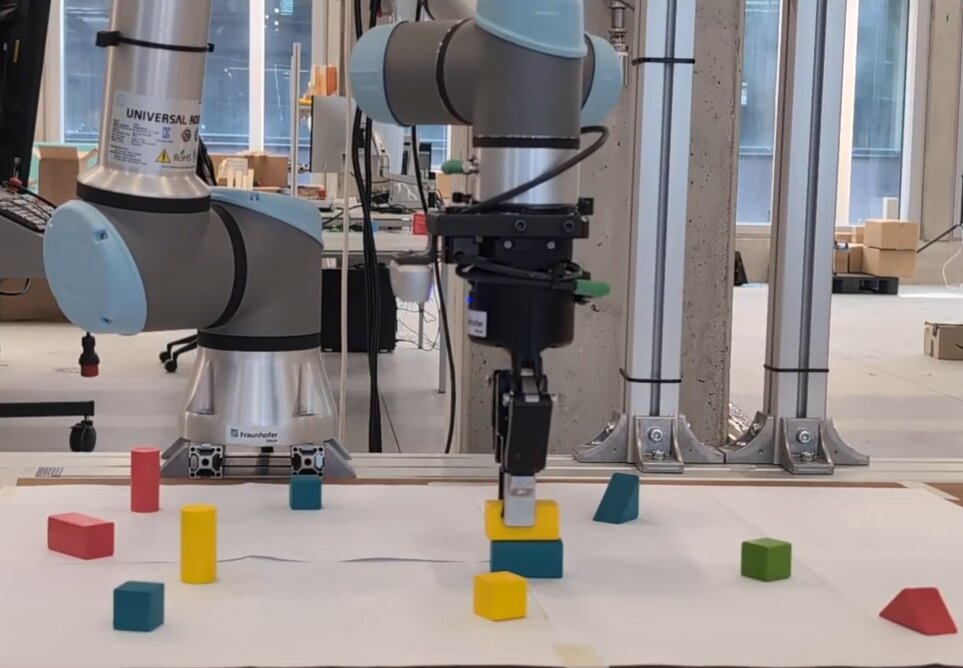}
    \includegraphics[width=0.24\linewidth]{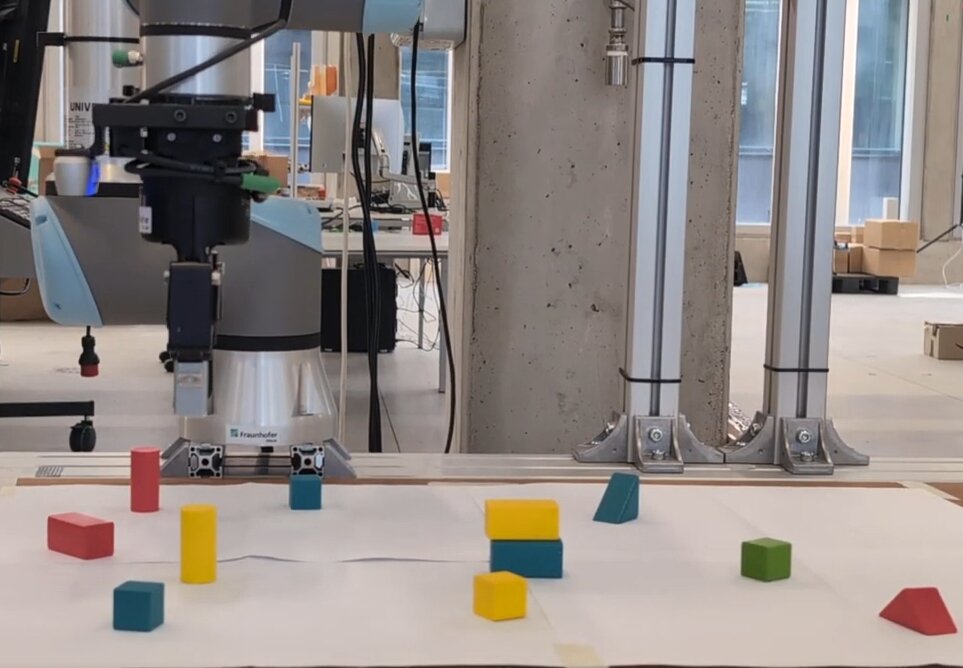}
    \caption{Execution of the BT reported in Fig.\ref{fig: example BT print}.}
    \label{fig: execute printed BT}
\end{figure}

\subsection{Discussion of Experimental Results}
The experimental results provide evidence that structured robot policies can be learned from fully synthetic supervision without relying on real-world demonstrations. Moreover, the results indicate that the model internalizes structural priors of robotic policies rather than memorizing fixed templates, as demonstrated by consistent syntactic validity and successful transfer across novel scene configurations and different robotic platforms. By bypassing the traditional bottlenecks associated with manual data collection and annotation in robotics, the proposed pipeline enables scalable generation of multimodal instruction–policy pairs while preserving the logical structure required by Behavior Trees.
Despite being trained exclusively on synthetic visual inputs and symbolic annotations, the fine-tuned vision-language model successfully internalizes the hierarchical grammar of robotic control policies and generalizes to real-world execution across different manipulators. This suggests that symbolic policy representations provide a stable interface between multimodal perception and robot control, facilitating transfer across heterogeneous robotic platforms.
The dual-platform validation on the Franka Emika Panda and UR5e further indicates that treating robotic primitives as symbolic abstractions supports logical consistency independently of the underlying hardware implementation. Overall, the results indicate that multimodal foundation models can be specialized to generate interpretable and executable symbolic policies, providing a complementary structured planning layer for modular robotic architectures. In fact, representing robot actions as symbolic primitives provides a hardware-agnostic policy interface, allowing the same learned structure to generalize across manipulators with different kinematic properties.

\section{Generalization and Cross-Domain Robustness} \label{Sec: Generalization}
The transition from high-fidelity synthetic environments to real-world interaction typically represents a significant bottleneck for robot learning methods. In this work, we posit that the adoption of a symbolic intermediate representation serves as a robust abstraction layer that facilitates seamless cross-domain transfer. This section investigates the \textbf{Invariance Robustness} of the fine-tuned Pixtral-12B model. We aim to demonstrate that the model does not merely perform pattern matching on synthetic MuJoCo pixels, but has instead internalized a universal spatial-logic prior. We evaluate this generalization across three distinct axes: semantic invariance to novel objects, deictic reasoning in ambiguous scenes, and hardware-agnostic logic transfer between heterogeneous robotic platforms.

\subsection{Visual Domain Invariance and Sim-to-Real Gap}
    A significant challenge in Sim-to-Real transfer is the discrepancies in lighting, texture, and background noise. While our training data (Section~\ref{sec:dataset_creation}) utilized Domain Randomization, it cannot exhaustively cover the visual complexity of a physical laboratory. However, we observed that the fine-tuned VLM exhibits high invariance to these low-level features. During real-world trials, the model successfully synthesized correct BTs despite the presence of overhead shadows, varying exposure levels, and background clutter (e.g., workstation equipment) that were entirely absent in simulation. We attribute this to the foundational visual priors of the base Pixtral model, which allow the specialized adapter to prioritize high-level geometric relationships over low-level pixel distributions.

\subsection{Visual Referential Grounding and Spatial Reasoning}\label{sec: visual referential}
    A critical question in VLM-based planning is whether the model utilizes the visual observation $I$ for reasoning or relies solely on the symbolic metadata $S$. To evaluate this, we designed a test case involving \textbf{referential ambiguity}. The scene contained two cylinders with different colors and other figures (left image in Fig.~\ref{fig: ablation images}): a \texttt{yellow\_cylinder\_0} positioned on the left and a \texttt{green\_cylinder\_1} on the right. The model was given the ambiguous instruction $U$: \textit{"Pick up the left cylinder"}, without specifying the color of the object. The model correctly identified \texttt{yellow\_cylinder\_0} as the target based on its relative spatial position in the RGB image.

    This result demonstrates that the fine-tuned Pixtral-12B effectively bridges the gap between high-level deictic expressions (relative spatial terms) and the discrete symbolic labels provided in the metadata. While the metadata provides the coordinates, the \textit{association} between the natural language "left" and the specific object label is resolved through visual reasoning. This confirms that the vision backbone is essential for grounding instructions in real-world environments where objects are defined by their relative context rather than unique symbolic names.

\subsection{Compositional Visual-Symbolic Grounding under Novel object geometries}

\begin{figure}[h]
    \centering
    \includegraphics[width=0.32\linewidth]{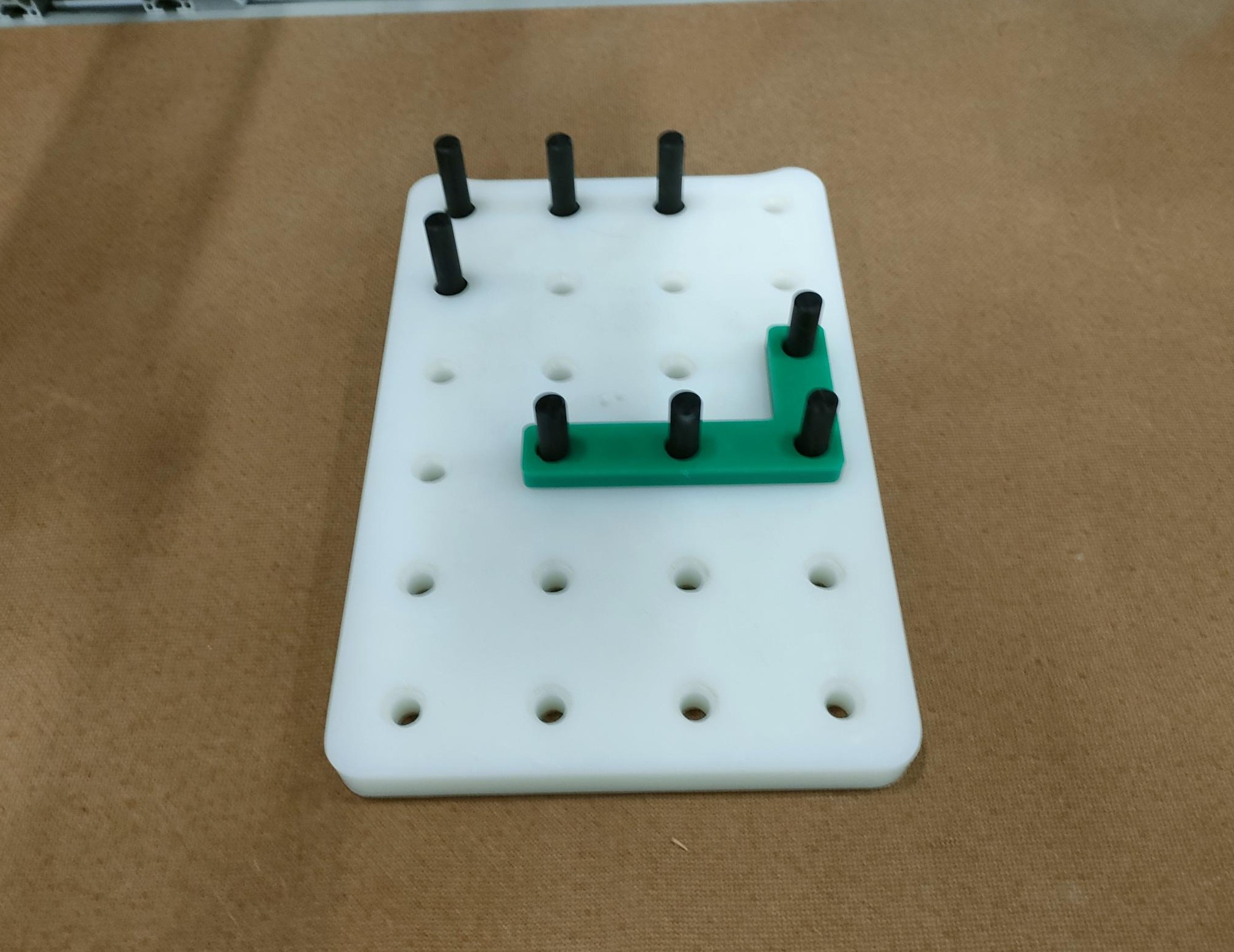}
    \includegraphics[width=0.32\linewidth]{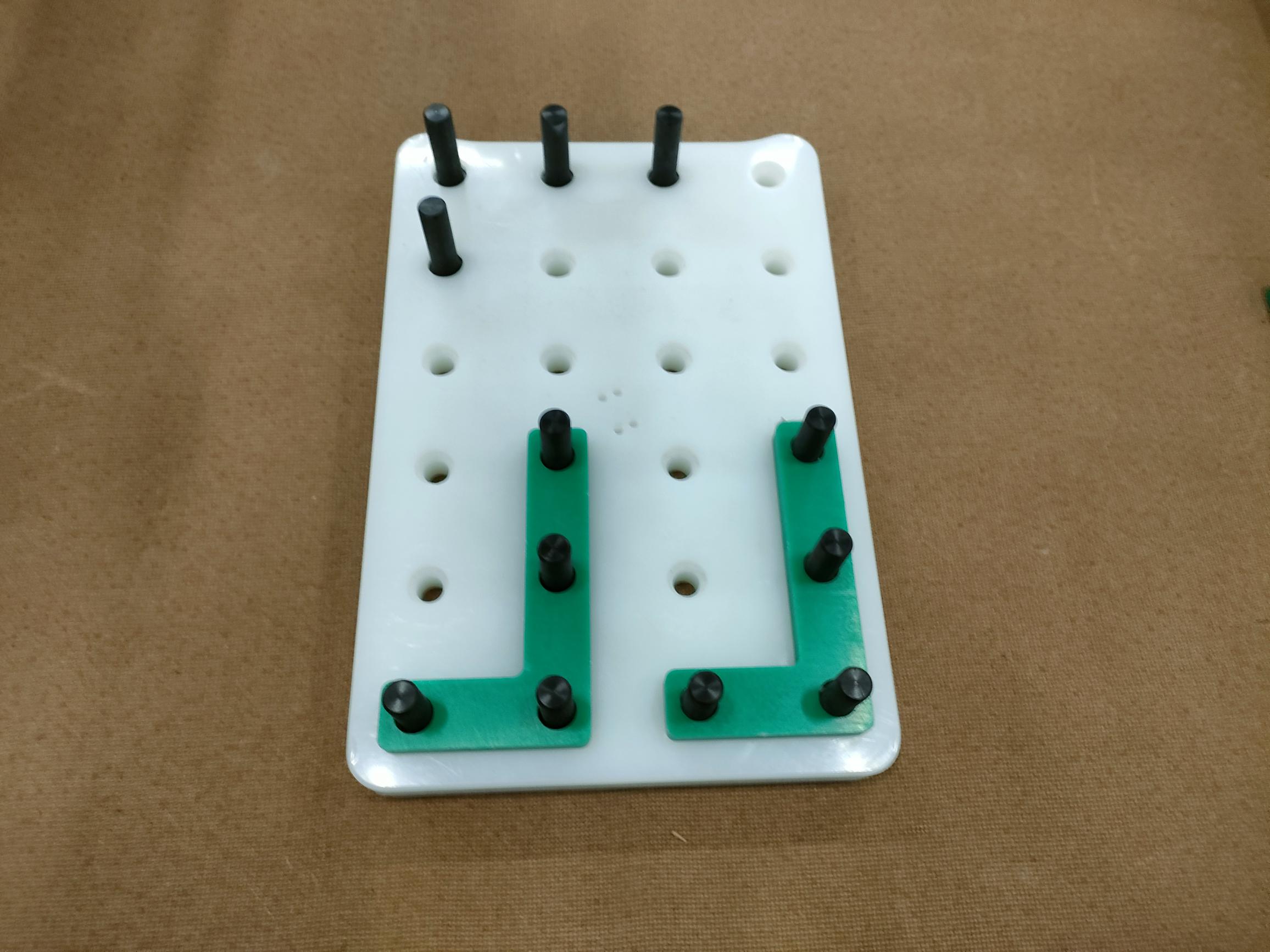}
    \includegraphics[width=0.32\linewidth]{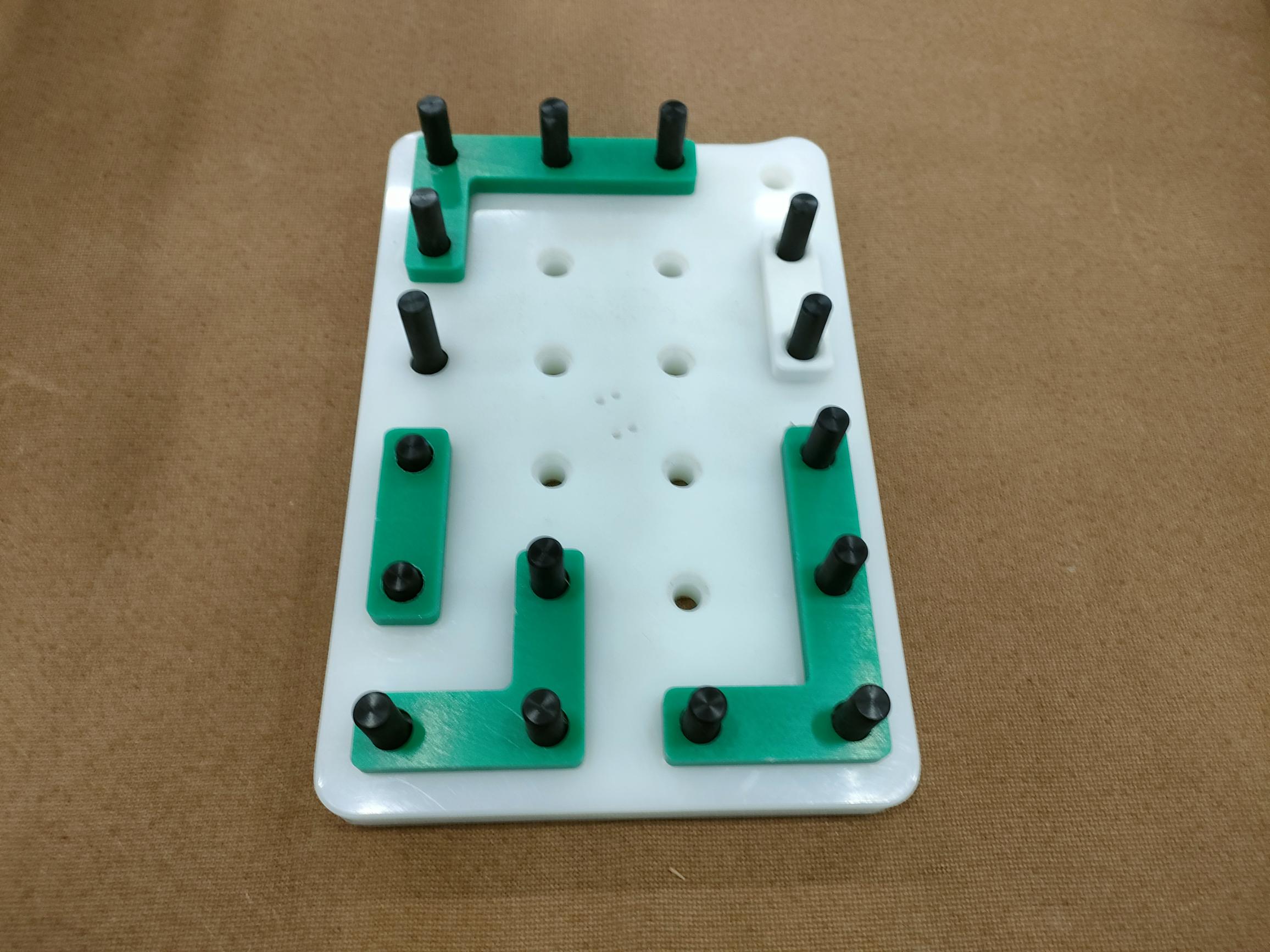}
    \caption{Validation on unseen components. The images show the setup for the zero-shot generalization test. The model successfully identifies the specific parts and their relative positions, translating high-level user commands into executable Behavior Trees for tasks.}
    \label{fig: new objs}
\end{figure}

A fundamental aspect in synthetic supervision is whether the model's perception is over-fitted to the simplified geometric primitives (e.g., prisms, cylinders) present in the training simulation. To test the boundaries of visual generalization, we conducted experiments using classes of objects absent from the training set: organic entities (e.g., various fruits), specialized industrial components, and other new unseen shapes as reported in Fig.~\ref{fig: new objs}.

\begin{figure}
    \centering
    \includegraphics[width=0.24\linewidth]{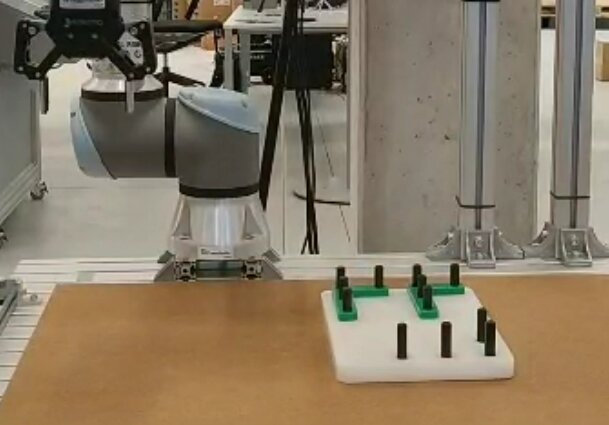}
    \includegraphics[width=0.24\linewidth]{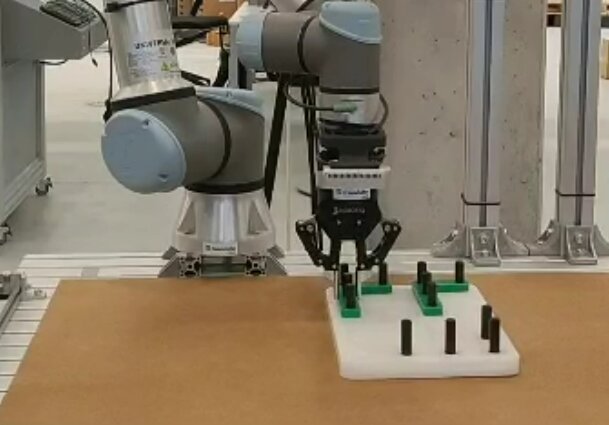}
    \includegraphics[width=0.24\linewidth]{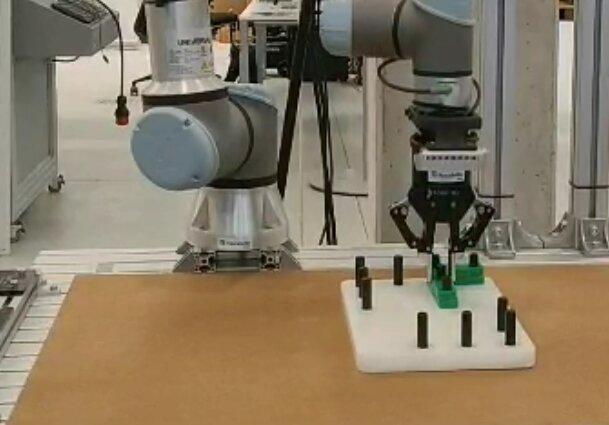}
    \includegraphics[width=0.24\linewidth]{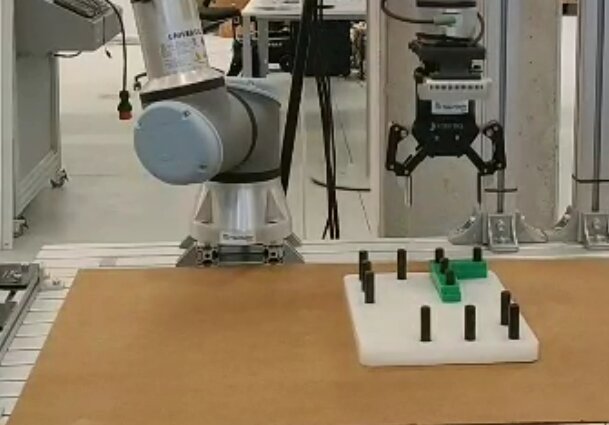}
    \caption{Execution in the real world of an in-distribution task with out-of-distribution and unseen objects.}
    \label{fig: real world exec}
\end{figure}

In this case, the model was presented with a real-world assembly task involving a white pegboard and multiple shaped green and white components, labeled in the metadata as \texttt{left\_green\_l\_1}, \texttt{top\_green\_l\_0}, etc. Despite the training set containing only primitive cylinders and boxes, the model correctly performed referential grounding on these unseen complex shapes. For example, when prompted to \textit{"Stack the L-shaped objects"}, the model successfully mapped the linguistic description to the high-dimensional visual features of the industrial part, correctly identifying its symbolic label from the metadata (Fig.~\ref{fig: real world exec}).

These results indicate that the fine-tuning process on 10,000 domain-randomized synthetic samples does not merely teach the model to recognize specific shapes. Instead, it appears to enhance the VLM's ability to \textbf{project natural language deictic expressions onto arbitrary visual clusters}. This semantic robustness is critical for industrial applications where the robot must interact with custom-designed parts or varying organic products that cannot be feasibly modeled in a simulation for every individual training run. The successful execution of synthesized BTs in these scenarios confirms that the framework achieves a high degree of \textbf{Visual-Symbolic Generalization}, bridging the gap between idealized simulation and the complexity of real-world objects.

\subsection{Cross-Platform Hardware Invariance}
    The successful execution of identical Behavior Trees on the Franka Emika Panda and the UR5e (Section~\ref{Sec: real-world}) highlights the portability of our neuro-symbolic approach. Because the VLM synthesizes logic over an abstract primitive library rather than low-level joint-space trajectories, the resulting policy is invariant to the specific kinematics of the manipulator. This confirms that the model has internalized a universal robotic task logic, enabling zero-shot deployment across a heterogeneous fleet.

\section{Boundary Analysis and Current Limitations} \label{Sec: Limitations}
The deployment of Large Multimodal Models in physical systems often reveals a discrepancy between semantic competence and functional execution. While the proposed framework demonstrates high reliability in grounded manipulation, it is equally critical to analyze the scenarios where the logic breaks down. We characterize these failure modes not as random stochastic errors, but as \textbf{Structural Boundaries} in zero-shot procedural synthesis. This section provides a systematic investigation into the limits of current VLM-based planning, specifically focusing on the gaps in procedural compositionality and the model’s lack of intrinsic causal reasoning about physical interaction.

%\section{Limitations and Generalization}
%While the previous sections established the framework's performance on in-distribution tasks, a core objective of neuro-symbolic robot learning is to achieve generalization beyond the specific parameters of the training environment. Traditional end-to-end models often suffer from a drop in performance when encountering out-of-distribution (OOD) visual features or complex semantic verbs. In this section, we investigate the boundaries of the fine-tuned Pixtral-12B model by exposing it to scenarios that challenge its visual perception, its logical depth, and its ability to adapt to novel hardware specifications. Specifically, we evaluate the zero-shot transferability of the learned spatial logic to objects with unseen geometries and the model's capacity to handle high-level semantic instructions that were not explicitly represented in the 10,000-sample synthetic dataset.

\subsection{Zero-Shot Generalization to Unseen Semantic Tasks}
    To evaluate the model's ability to generalize beyond the explicit instructions found in the synthetic training set, we introduced the high-level command $U$: \textit{"Switch the positions of the two cylinders"} and $I$, same as Sec.~\ref{sec: visual referential}. Unlike simple pick-and-place tasks, a "switch" operation is an \textbf{unseen semantic task} that requires the model to perform long-horizon planning and utilize intermediate workspace management. 
    Specifically, for a single-arm manipulator to swap the locations of two objects, it must autonomously synthesize a sequence that involves:  Moving the first object to a temporary buffer (\texttt{temp\_pose}), moving the second object to the vacated original pose of the first, and moving the first object from the buffer to the original pose of the second. This test evaluates whether the fine-tuned VLM can decompose a novel high-level verb into a multi-stage execution chain using the atomic primitives defined in Section~\ref{Sec: Atomic Actions}. 
    
    The output reveals a significant behavior we characterize as \textbf{Pattern Backsliding}. While the model successfully identified both \texttt{green\_cylinder\_1} and \texttt{yellow\_cylinder\_0} and resolved their spatial relationship, the resulting plan did not achieve a swap. Instead, the model synthesized a sequence to pick up the green cylinder and place it directly at the approach pose of the yellow cylinder, effectively attempting to stack them or place them in the same pose.
    %A successful synthesis of a "Switch" BT—despite the absence of "switch" examples in the 10,000 training samples—would indicate that the model has not merely memorized templates but has instead learned a functional mapping between spatial goals and symbolic logic.
    This failure identifies a clear boundary in the 12B-parameter model's zero-shot capabilities. Although the model exhibits high spatial grounding, it lacks the logical depth to autonomously decompose "Switch" into a buffer-based sequence without explicit demonstrations. For researchers, this suggests that while fine-tuning on synthetic data is highly effective for mastering a domain's grammar and grounding, complex multi-stage procedural reasoning still requires either a more diverse training distribution or an iterative planning loop to correct sub-optimal logic. This result highlights that compositional generalization in symbolic policy generation remains sensitive to the diversity of semantic task structures observed during training.
    The framework, in fact, focuses on structured policy synthesis rather than general symbolic reasoning, and performance may depend on coverage of semantic task patterns within the synthetic supervision.

\subsection{Zero-Shot Generalization to Out-of-Vocabulary Primitives}    
    To evaluate the extent to which the model treats the System Specification ($S$) as a dynamic source of truth, we performed a zero-shot end-effector transfer experiment. Although the 10,000-sample synthetic training set exclusively utilized parallel gripper primitives (e.g., \texttt{OpenGripper}), we modified $S$ at inference time to describe a vacuum-based suction system. This system introduced entirely out-of-vocabulary primitives: \texttt{EngageVacuum}, \texttt{ReleaseVacuum}, and the status check \texttt{is\_suction\_seal}.

    The task instruction $U$ remained coherent with the training dataset: \textit{"Stack the yellow cylinder on the green cylinder"} (using left image in Fig~\ref{fig: ablation images}). This test serves as a benchmark for \textbf{cross-hardware functional generalization}. 
    
    The resulting Behavior Tree reveals a phenomenon we define as the \textbf{Functional Grounding Gap}: the model successfully generalizes lexical vocabulary but fails to generalize functional logic. The model demonstrated high lexical flexibility by correctly adopting the new tokens from $S$. It successfully replaced the "grasp" and "release" concepts with \texttt{EngageVacuum} and \texttt{ReleaseVacuum}, respectively. However, the model exhibited a critical failure in \textbf{Causal Grounding}. It attempted to execute the \texttt{EngageVacuum} selector at the very root of the tree, while the manipulator was still at the \texttt{home} pose. This suggests that while the VLM understands the name of the tool, it lacks a zero-shot understanding of the physics of interaction. Specifically, that suction requires contact before engagement. These results define a clear boundary for LLM-based task planning: spatial grounding and syntactic JSON compliance transfer easily, but tool-specific interaction logic remains partially in to the model pretrained-backbone. This proves that achieving true hardware-agnosticism requires a training distribution that spans multiple end-effector modalities, preventing the model from over-fitting to the mechanical constraints of a single paradigm.

\subsection{The Verification Gap and Formal Guarantees}
    A significant challenge in current neuro-symbolic robotics is the lack of a generalized framework for the formal verification of Large Multimodal Model (LMM) outputs. In this work, the synthesized policies were manually checked to ensure alignment with the task execution goal. This dependency on human oversight defines the \textbf{Verification Gap}: while the model is capable of generating highly complex and syntactically correct trees, it cannot yet provide a formal proof of safety for its own generated logic. 

    This manual dependency is not a failure of the model's generative capacity, but rather a reflection of the current absence of automated symbolic checkers that can translate high-dimensional visual context into formal robotic invariants. Future research will explore the integration of formal methods, such as Linear Temporal Logic (LTL) checking~\cite{LTL1, LTL2}, to replace the human-in-the-loop stage with a provably safe automated filter.

\subsection{Lexical Coherence and Symbolic Hallucination}
To evaluate the framework's robustness against adversarial users, we conducted tests in which the model was instructed to manipulate objects that were completely absent from the visual scene $\mathcal{I}$ and symbolic metadata in $\mathcal{S}$ (e.g. \textit{"Put the black object on top of the white object"} in a scene containing neither, without even specifying the shape). The model's response revealed boundaries in open-loop VLM planning with symbolic hallucinations. Rather than failing to generate the JSON that accomplishes the user request, the model synthesized a grammatically flawless and executable BT. The model hallucinated arguments that perfectly mimic the naming convention of the training metadata, generating labels such as \texttt{black\_cylinder\_0} and \texttt{white\_cube\_1} and successfully applied the \texttt{z\_offset} arithmetic convention.

This behavior yields two significant insights. First, while base foundation models inherently suffer from hallucination when prompted with out-of-context objects (non-trained Pixral12B suffers from the same hallucination with textual output), our supervised fine-tuning ensures that this failure mode is strictly channeled into syntactically valid JSON-BT grammar rather than catastrophic formatting degradation. 
Second, and most importantly, this highlights the inherent safety advantage of utilizing Behavior Trees over end-to-end motor policies. In a traditional black-box Vision-Language-Action (VLA) model, a visual hallucination could trigger unpredictable and potentially dangerous joint-space trajectories. In our neuro-symbolic framework, the hallucinated execution plan is intrinsically safe: during physical deployment, the very first condition check guarding the manipulation (e.g., \texttt{is\_at\_pose(black\_cylinder\_0)}) will simply evaluate to \texttt{FAILURE} because the external perception module will not locate the hallucinated symbol. Consequently, the tree gracefully halts without executing any motor actions. This demonstrates that structured policies effectively encapsulate VLM hallucinations within verifiable execution bounds, decoupling semantic reasoning errors from physical hardware safety.

Our boundary analysis reveals that while synthetic supervision effectively teaches the syntactic grammar of physical interaction (e.g., pick, place), the model lacks intrinsic zero-shot physics simulators in its weights. Therefore, generalizing to unseen semantic verbs (like 'Switch') or novel mechanics (like 'Suction') requires representation in the training distribution. Because our synthetic data generation is highly scalable, future work can seamlessly inject tool-use or multi-stage tasks into the simulation engine to close this causal gap.

\section{Conclusions and Future Works}\label{Sec: conclusion}
This work investigated how vision-language models can be specialized to generate structured robot policies grounded in multimodal perception. By combining neural perception with symbolic policy representations, the proposed framework enables the synthesis of executable Behavior Trees that preserve interpretability, modularity, and reactive execution.
To enable scalable supervision without manual annotation, we introduced a fully synthetic training pipeline that generates multimodal instruction–policy pairs, allowing the model to internalize structural and spatial reasoning constraints required for symbolic policy generation.
Experimental results show that structured policies learned entirely from synthetic supervision transfer successfully to real-world robotic platforms, maintaining logical consistency across different manipulators. These findings indicate that multimodal foundation models can be adapted to produce interpretable symbolic policies, providing a structured alternative to opaque end-to-end visuomotor approaches.
Future work will explore more complex task compositions and tighter integration between symbolic policy synthesis and low-level control.

\bibliographystyle{IEEEtran}
\bibliography{references}

\end{document}